\documentclass{article}

\usepackage{arxiv}

\usepackage[utf8]{inputenc} % allow utf-8 input
\usepackage[T1]{fontenc}    % use 8-bit T1 fonts
\usepackage{hyperref}       % hyperlinks
\usepackage{url}            % simple URL typesetting
\usepackage{booktabs}       % professional-quality tables
\usepackage{amsfonts}       % blackboard math symbols
\usepackage{nicefrac}       % compact symbols for 1/2, etc.
\usepackage{microtype}      % microtypography
\usepackage{lipsum}
\usepackage{graphicx}
\usepackage{listings}
\usepackage{mdframed}
\usepackage{varwidth}
\usepackage{subcaption}
\usepackage[style=ieee, backend=biber]{biblatex}
\usepackage[acronym]{glossaries}
\usepackage[mathscr]{euscript}
\usepackage{mathtools}
\usepackage[toc, page]{appendix}
\usepackage{xpatch}
\usepackage{float}
\usepackage{multirow}
\usepackage{booktabs}
\usepackage{xcolor}
\usepackage{subcaption}
\usepackage{authblk}

\hypersetup{colorlinks=true,urlcolor={blue!40}}
\newcommand{\deemph}[1]{{\color{black!40}#1}}

\graphicspath{ {./images/} }

\title{OpenBezoar: Small, Cost-Effective and Open Models Trained on Mixes of Instruction Data}

\author{\textbf{Chandeepa Dissanayake}}
\author{\textbf{Lahiru Lowe}}
\author{\textbf{Sachith Gunasekara}}
\author{\textbf{Yasiru Ratnayake}}
\affil{Surge Global}
\affil{\textit{\{chandeepa, lahiru.lowe, sachith, yasiru\}@surge.global}}

\begin{document}

\newacronym{acr_llm}{LLM}{Large Language Model}
\newacronym{acr_llms}{LLMs}{Large Language Models}
\newacronym{acr_sft}{SFT}{Supervised Fine-Tuning}
\newacronym{acr_dpo}{DPO}{Direct Preference Optimization}
\newacronym{acr_lora}{LoRA}{Low Rank Adaptation}
\newacronym{acr_qlora}{QLoRA}{Quantized Low Rank Adaptation}
\newacronym{acr_rl}{RL}{Reinforcement Learning}
\newacronym{acr_rlhf}{RLHF}{Reinforcement Learning from Human Feedback}
\newacronym{acr_ppo}{PPO}{Proximal Policy Optimization}

\maketitle
\begin{abstract}
  Instruction fine-tuning pretrained \acrshort{acr_llm}s for diverse downstream tasks has demonstrated remarkable success and has captured the interest of both academics and practitioners. To ensure such fine-tuned \acrshort{acr_llm}s align with human preferences, techniques such as RLHF and \acrshort{acr_dpo} have emerged. At the same time, there is increasing interest in smaller parameter counts for models. In this work, using OpenLLaMA 3Bv2 as a base model, we describe the recipe used to fine-tune the OpenBezoar family of models. In this recipe: We first generate synthetic instruction fine-tuning data using an open and commercially non-restrictive instruction fine-tuned variant of the Falcon-40B model under three schemes based on: LaMini-LM, WizardLM/Evol-Instruct (with databricks-dolly-15k as a seed dataset) and Orca (with the Flan Collection as a seed dataset), then filter these generations using GPT-4 as a human proxy. We then perform cost-effective \acrshort{acr_qlora}-based supervised fine-tuning sequentially with each scheme. The resulting checkpoint is further fine-tuned with a subset of the HH-RLHF dataset to minimize distribution shift prior to using the \acrshort{acr_dpo} loss to obtain the final checkpoint. Evaluation is done with the LM Eval Harness tasks/metrics as well as on MT-Bench using the ``LLM-as-a-judge'' framework with Claude 2.1, with the finding that the final checkpoint, ``OpenBezoar-HH-RLHF-DPO'', demonstrates superior performance over many models at the 3B parameter scale, even outperforming the top model in one of the categories on the Huggingface Open LLM Leaderboard. We release ``OpenBezoar-SFT'', ``OpenBezoar-HH-RLHF-SFT'', ``OpenBezoar-HH-RLHF-DPO'' checkpoints, alongside our generated datasets on HuggingFace \href{https://huggingface.co/collections/SurgeGlobal/open-bezoar-6620a24923e12127e9e2b9cc}{here} and our codebases \href{https://bitbucket.org/paladinanalytics/workspace/projects/OP}{here}.\par
\end{abstract}

% keywords can be removed

\section{Introduction}
%% First describe all other stuff and come down here.

\acrfull{acr_sft} of pre-trained \acrfull{acr_llms} on instruction datasets in order to specialize them in a variety of downstream tasks is not just pivotal for guiding them to produce sensible responses, but also serves as a compelling demonstration of how supervised learning can enable artificial models to generalize effectively through observational learning. This process of \acrshort{acr_sft} for \acrshort{acr_llm}s is largely similar to other gradient based optimization pipelines. Early examples, and at present the most capable \acrshort{acr_llm}s are very large, with reported parameter count exceeding 100B. Consequently, the computational cost of \acrshort{acr_sft} for an \acrshort{acr_llm} of such magnitude is out of reach for organizations and individuals with conventional budgetary expectations. However, it has been demonstrated that models with a comparatively smaller parameter count can perform reasonably well on diverse downstream tasks, even outperforming larger models in specific cases\cite{touvron2023llama} \cite{gunasekar2023textbooks}. Our survey during the preliminary stages of this work (carried out between March and October of 2023) indicated that there were only a handful of fine tuned models at the 3B scale that had benchmark scores comparable to their larger counterparts\cite{togethercomputer_RedPajama_INCITE_Base_3B_v1}. Subsequently, in an effort to investigate the potential of instruction fine-tuned 3B parameter scale models, we chose to devise and implement a recipe for fine-tuning the OpenLLaMA 3B V2\cite{openlm2023openllama} base model, which was a very recently released model at the time.\par

The largest and most capable base models are typically fine-tuned using datasets comprising of large numbers of human-generated examples, which in part accounts for the versatility of the resulting fine-tuned \acrshort{acr_llms}. However, such datasets are costly to create due to the human labor required, cause training times to inflate, and typically, their resulting models' licenses prevent the commercial use of novel models fine-tuned on such models' outputs. Open, crow-sourced datasets exist, but they often suffer from problems such as limited diversity and relatively small size\cite{DatabricksBlog2023DollyV2,together2023redpajama, openassistantoaast1}. Scale can be achieved by having an \acrshort{acr_llm} generate completely new instruction datasets \cite{maeng2019alpaca, wang-etal-2023-self-instruct}, but the most capable such models have restrictive licensing, casting uncertainty on the openness of derived models trained on their outputs. Our aim in this work is to utilize a sufficiently capable open-source instruction model with a license that permits commercial use of the generated responses \cite{h2oaifalcon40boasst}, in order to generate instruction/response pairs via three dataset generation schemes, resulting in instruction datasets that permit commercial use. We go on to further filter this dataset for higher quality and more diverse generations using a better human proxy model \cite{achiam2023gpt}, and perform \acrshort{acr_sft} on our chosen open base model using \acrshort{acr_qlora}, resulting in three \acrshort{acr_qlora} adapter models. These models, along with an alignment-specific model described below, comprise the OpenBezoar family of models, released herewith.\par

As outlined above, \acrshort{acr_llms} when fine-tuned using supervised methods for different tasks on large datasets have been proven to generalize surprisingly well and perform on a wide range of benchmarks. If these acquired expertise are collectively termed the model's ``skillset"\footnote{This may be taken figuratively rather than literally, as the latter falls back to the question: ``Are \acrshort{acr_llm}s truly intelligent?" \cite{bubeck2023sparks}}, some of these skills might not be desirable under certain scenarios or may need to be modulated subject to certain circumstances. For instance, if asked to generate a plan to conduct a criminal activity after providing a detailed background of the target of interest, a naively fine-tuned \acrshort{acr_llm} might choose to respond back with the guidelines to achieve the task. While this is to be expected, a human in similar circumstances may elect to exercise more agency and question the request or refuse to answer it. It may be prudent to endow models with similar capabilities dependent on context. To further anthropomorphize, certain responses might be preferred over others based on the context as well. This leads to the conclusion that it may be advisable to \textit{bias} the output of the \acrshort{acr_llm} towards the human-preferred output, which can be achieved through further fine-tuning the \acrshort{acr_llm} with an objective that achieves the \textbf{alignment} desired.\par

As the generation using \acrshort{acr_llms} is discrete by nature (as it proceeds token-wise), the objective function for such alignment fine-tuning is inherently non-differentiable. Consequently, a popular approach is to optimize weights post-hoc using \acrfull{acr_rl}, called \acrfull{acr_rlhf} in this context. More specifically, to maximize a reward based on human preference using \acrfull{acr_ppo}\cite{ziegler2020finetuning} is now common. The objective for \acrshort{acr_rl} necessitates a reward model that has been trained on a comparisons dataset sampled from a preference distribution, modelled with a preference model such as Bradley-Terry\cite{Bradley1952RankAO}. A prerequisite for implementing preference modeling techniques is that human annotators, either online or more commonly offline, label the answers to prompts with a ranking that denotes their preferences. As fine-tuning \acrshort{acr_llms} is typically orchestrated at a large scale on massive datasets, the requirement to separately train a reward model can be a significant bottleneck due to these requirements. However, more recently, it has been shown that with a change of variables it is possible to express the objective for training a reward model as a function of the policy itself\cite{rafailov2023direct}, allowing us to dispense with the reward model and make reward implicit. This technique of \acrfull{acr_dpo} allows the alignment of \acrshort{acr_llms} directly from preference datasets. In our work, we perform \acrshort{acr_dpo} on a subset of the Anthropic HH-RLHF dataset\cite{bai2022training} after merging the \acrshort{acr_qlora} adapter from the \acrshort{acr_sft} stage. We deliberately chose to apply \acrshort{acr_dpo} to the merged model as the update rule of \acrshort{acr_dpo} explicitly refers to the entire parameterized \acrshort{acr_llm}\cite{rafailov2023direct}. Further research is required to evaluate the use of low-rank adapters in this regard.\par

We release checkpoints after each stage of \acrshort{acr_sft} and the merged models before\footnote{Checkpoint after performing \acrshort{acr_sft} on a subset of HH-RLHF dataset to minimize the distribution shift} and after \acrshort{acr_dpo}. We call the merged model after the final \acrshort{acr_sft} checkpoint ``OpenBezoar-SFT" and models before and after \acrshort{acr_dpo}, ``OpenBezoar-HH-RLHF-SFT", and ``OpenBezoar-HH-RLHF-DPO" respectively. Out of ten benchmarks evaluated, OpenBezoar-SFT outperformed the base model in all but two benchmarks (SciQ, PIQA), significantly outperforming it on TruthfulQA (14.18\% accuracy improvement), OpenBookQA (8.84\%), and MMLU (4.29\%), with an overall average improvement of 1.48\%. The final OpenBezoar-HH-RLHF-DPO model outperforms OpenBezoar-SFT in turn on average by 2.36\%, recording improvements on all benchmarks except TruthfulQA (-2.75\%) and MMLU (-6.04\%).
In order to evaluate human preferences alignment, we employ the \acrshort{acr_llm}-as-a-judge framework\cite{zheng2023judging} with the MT-bench benchmark question set. Although it has been established that GPT-4 matches human preferences by achieving the same level of agreement as among humans, here we attempt to establish Anthropic's Claude-2.1\cite{anthropic} as a viable judge. In this regard, we calculate the agreement between Claude-2.1 and other types of judges, including humans and observe that Claude-2.1 exceeds the threshold of 80\% agreement, thus validating its potency as a judge to approximate human preferences. Subsequently, we first evaluate OpenBezoar, OpenBezoar-HH-RLHF-SFT, and OpenBezoar-HH-RLHF-DPO models for single answer grading\footnote{Refer to \cite{zheng2023judging} for other variants} to establish the overall dominance of OpenBezoar-HH-RLHF-DPO over the preceding models. Hence we choose OpenBezoar-HH-RLHF-DPO for evaluations against three other publicly available models, only one of which (RedPajama-INCITE-Chat-3B-v1) was available at the time of experimentation\footnote{Our experiments were primarily done over June-September of 2023.}. These models were chosen based their having a comparable parameter count and their ranking in the HuggingFace Open \acrshort{acr_llm} Leaderboard\cite{huggingface-open-llm-leaderboard}. In terms of the average score, OpenBezoar-HH-RLHF-DPO model surpassed two out of the three competitors, and even outperformed the top performing chat model in 3B parameter scale in one of the categories (Writing).\par

\section{Preliminaries}
\subsection{Dataset Creation}
\label{preliminaries:dataset_creation}
\subsubsection{LaMini}

The LaMini approach developed by \Citeauthor*{wu2023lamini} \cite{wu2023lamini} involves generating a large-scale instruction dataset by leveraging the outputs of a large language model, gpt-3.5-turbo. The authors use two strategies for generating instructions: example-guided and topic-guided. The example-guided strategy involves providing a few seed examples and constraints to gpt-3.5-turbo and asking it to generate diverse instructions that follow the same format and style. The topic-guided strategy involves using common topics collected from Wikipedia to guide the generation process and expand the scope of the instructions. The authors then use gpt-3.5-turbo to generate responses for each instruction, resulting in a dataset of 2.58 million instruction-response pairs.

\subsubsection{Evol-Instruct pipeline}

\citeauthor{xu2023wizardlm} \cite{xu2023wizardlm} propose the Evol-Instruct pipeline for automatically evolving
instruction datasets using large language models (LLMs). Starting from an initial dataset $D^{(0)}$,
Evol-Instruct iteratively upgrades the instructions in each evolution step to obtain a sequence of
evolved datasets $[D^{(1)} \dots D^{(M)}]$. The pipeline consists of two main components: 1) an
Instruction Evolver that leverages an LLM with specialized prompts to perform in-depth evolving,
which increases the complexity of instructions, and in-breadth evolving, which enhances the
diversity of instruction topics and skills; and 2) an Instruction Eliminator that filters out
unsuccessfully evolved instructions based on criteria such as lack of information gain or the LLM's
inability to generate a meaningful response. By alternating between these evolving and eliminating
steps, Evol-Instruct produces an increasingly rich and challenging instruction dataset.

\subsubsection{Orca}

The Orca approach \cite{mukherjee2023orca} aims to overcome the limitation of imitation learning, whereby smaller models trained on outputs of a LFM (large foundation model) tend to learn to imitate the style, but not the reasoning process of the LFM in question. Orca on the other hand leverages explanation tuning, where ⟨query, response⟩  pairs of vanilla instruction tuning methods are augmented by detailed responses generated from GPT-4. This system acts as a teacher - student mechanism where the LFM acts as a teacher from which the detailed responses are generated and the student, i.e. the smaller learns from. These detailed responses are elicited by 16 different system messages providing an opportunity for smaller models to mimic the “thought” and “reasoning” process of a LFM. These system instructions could also be used as a safety harness for improving the safety of the smaller models' responses.

\subsection{Human Preferences Alignment}
\label{preliminaries:human_preferences}
\acrshort{acr_llm}s are pre-trained with the simple language modelling objective of predicting the next token. Thus, the outputs of the \acrshort{acr_llm}s are susceptible to unintended behaviours such as hallucinations, high degree of toxicity, and factual inconsistencies. For any purpose other than the unguided creativity, these \textit{misaligned} \acrshort{acr_llm}s should be aligned with additional and supplemental objectives. Such new objectives might often be conflicting with the objectives that the \acrshort{acr_llm} has already been fine-tuned upon. For example, while we may require the outputs to precisely follow instructions, it might also be demanded that it should be harmless. The typical approach is to perform \acrfull{acr_rlhf} for implicit objectives\footnote{According to \cite{askell2021general}, helpfulness, honesty and harmlessness are demanded} on \acrshort{acr_llm}s that has been fine-tuned for relevant downstream tasks\cite{ouyang2022training}.\par

\subsubsection{\acrshort{acr_rlhf} Pipeline}
The typical \acrshort{acr_rlhf} pipeline includes three phases.
\begin{enumerate}
    \item \acrshort{acr_sft} on a pre-trained \acrshort{acr_llm} with an appropriate dataset to obtain the fine-tuned \acrshort{acr_llm} $\pi^{SFT}$.
    \item Reward Modelling Phase
    \item Fine-Tuning with \acrshort{acr_rl}
\end{enumerate}

In the reward modelling phase, $\pi^{SFT}$ is presented with the prompt $x$ to generate two responses $y_1$ and $y_2$. Then the human annotators are tasked with ranking them, in this case as preferred($y_{w}$) and dispreferred($y_{l}$). The preference of $y_{w}$ over $y_{l}$ is denoted by $y_{w} \succ y_{l}$. Preferably, the Bradley-Terry model\cite{Bradley1952RankAO} is utilized to describe the distribution of such human preferences. Assuming a latent but inaccessible reward model $r^{*}(x,y)$ where $y$ is the response generated by $\pi^{SFT}$ to the given prompt $x$, the Bradley-Terry model postulates that the human preference distribution $p^{*}$ is given by,
\begin{equation}\label{eq:rlhf_bt_1}
    p^{*}(y_1 \succ y_2 \mid x) = \frac{\exp(r^{*}(x,y_1))}{\exp(r^{*}(x,y_1)) + \exp(r^{*}(x,y_2))}
\end{equation}\par

Accordingly, the dataset $\mathscr{D} = {(x^{(i)}, y_{w}^{(i)}, y_{l}^{(i)})}_{i = 1}^{N}$ that evolves through prompts $x$ and generating responses for them through $\pi^{SFT}$, can be considered to be sampled from $p^{*}$. Thus, a reward model can be parameterized as $r_{\phi}(x,y)$ to represent this dataset $D$ and the parameterized model can be estimated with the maximum likelihood method. Evidently, this resembles a binary classification problem, and therefore, the following negative log-likelihood loss can be used for reformulating this as an optimization problem.
\begin{equation} \label{eq:rlhf_rl_loss}
    \mathscr{L}_{RLHF}(r_{\phi}, \mathscr{D}) = - \mathbb{E}_{(x, y_w, y_l) \sim \mathscr{D}}\left[\log{\sigma(r_{\phi}(x, y_w) - r_{\phi}(x, y_l))}\right]
\end{equation}
where, $\sigma$ is the logistic function. $\mathscr{L}_{RLHF}$ is used to train the parameterized reward model $r_{\phi}$.\par

It is also worthwhile to note that if there are more than two responses and preference pairs between them, more general models such as Plackett-Luce ranking models\cite{luce_individual_nodate, plackett_analysis_1975} can be used to arrive at a similar result. Nevertheless, during the \acrshort{acr_rl} fine-tuning phase, $\pi^{SFT}$ is optimized using the \acrfull{acr_ppo} algorithm\cite{schulman2017proximal} for a KL-constrained reward maximization objective\cite{ziegler2020finetuning}.

\subsubsection{\acrlong{acr_dpo}}
The optimal solution($\pi_{r}$) to the KL-constrained objective of the fine-tuning phase of the \acrshort{acr_rlhf} pipeline can be given in the following form\cite{rafailov2023direct}. For any reward model $r$,
\begin{equation} \label{eq:dpo_optimal_sol}
    \pi_{r}(y \mid x) = \frac{1}{Z(x)} \pi_{\textnormal{ref}}(y \mid x) \exp{\frac{1}{\beta}r(x,y)}
\end{equation}
where $\beta$ is the hyperparameter in the fine-tuning phase of the \acrshort{acr_rlhf} pipeline which controls divergence from the base reference policy $\pi_{\textnormal{ref}}$, which is set to the initial $\pi^{SFT}$ before training in \acrshort{acr_rlhf} and thereby in \acrshort{acr_dpo}, and $Z$ is the partition function given by,
\begin{equation} \label{eq:dpo_optimal_sol_part_eq}
    Z(x) = \sum_{y}\pi_{\textnormal{ref}}(y \mid x) \exp{\frac{1}{\beta} r(x,y)}
\end{equation}
However, estimating $Z$ is expensive, rendering any direct utilization of the optimal solution in the Equation \ref{eq:dpo_optimal_sol} computationally worthless.\par

Rearranging Equation \ref{eq:dpo_optimal_sol} to express the reward model $r$ in terms of the optimal policy $\pi_{r}$ we obtain,
\begin{equation} \label{eq:dpo_rm}
    r(x,y) = \beta \log{\frac{\pi_{r}(y \mid x)}{\pi_{\textnormal{ref}}(y \mid x)}} + \beta \log{Z(x)}
\end{equation}
Since $r$ generally represents any reward model, setting $r = r^{*}$, substituting in the Equation \ref{eq:rlhf_bt_1} and finally with some algebra we get,
\begin{equation} \label{eq:dpo_bt}
    p^{*}(y_1 \succ y_2 \mid x) = \frac{1}{1 + \exp{\left(\beta \log{\frac{\pi^{*}(y_2 \mid x)}{\pi_{\textnormal{ref}}(y_2 \mid x)}} - \beta \log{\frac{\pi^{*}(y_1 \mid x)}{\pi_{\textnormal{ref}}(y_1 \mid x)}}\right)}}
\end{equation}
where $\pi^{*}$ is the optimal policy that corresponds to the latent reward model $r^{*}$.\par

Equation \ref{eq:dpo_bt} describes the human preferences distribution in terms of the optimal policy. Analogous to reward model parameterization in \acrshort{acr_rlhf}, the policy can now be parameterized and denoted by $\pi_{\theta}$, which corresponds to the parameterized reward model $r_{\phi}$. Therefore, using a similar substitution that we used to derive Equation \ref{eq:dpo_bt}, in the Equation \ref{eq:rlhf_rl_loss}, and by replacing $\mathscr{L}_{RLHF}(r_{\phi}, \mathscr{D})$ by $\mathscr{L}_{DPO}(\pi_{\theta}; \pi_{\textnormal{ref}})$ we get
\begin{equation} \label{eq:dpo_loss}
    \mathscr{L}_{DPO}(\pi_{\theta}; \pi_{\textnormal{ref}}) = - \mathbb{E}_{(x, y_w, y_l) \sim \mathscr{D}}\left[\log{\sigma\left(\beta \log{\frac{\pi_{\theta}(y_w \mid x)}{\pi_{\textnormal{ref}}(y_w \mid x)}} - \beta \log{\frac{\pi_{\theta}(y_l \mid x)}{\pi_{\textnormal{ref}}(y_l \mid x)}}\right)}\right]
\end{equation}
Instead of fitting the reward model with the loss in Equation \ref{eq:dpo_loss}, we directly fit the parameterization $\pi_{\theta}$. Most importantly, it is differentiable and therefore any vanilla optimizer can be used, thus eliminating the need for non-differentiable policy optimization.\par

Moreover, it is evident that the reward model is implicitly defined in this loss by the language model. Consequently, \acrshort{acr_dpo} eliminates the need for explicit reward modelling in \acrshort{acr_rlhf}. Additionally, in the \acrshort{acr_dpo} update, each training example is weighted by the difference between the implicit reward for dispreferred response and that of the preferred response, scaled by $\beta$\cite{rafailov2023direct}. Therefore, choosing the value for $\beta$ is trivial when preventing the language model from degenerating.\par

\paragraph*{\acrshort{acr_dpo} Pipeline}
Our emphasis is directed towards preference alignment through the utilization of an existing dataset. Phases in the \acrshort{acr_dpo} pipeline are outlined as follows.
\begin{enumerate}
    \item Perform \acrshort{acr_sft} on the language model, using the preferred responses of the dataset to ensure that $\pi_{\textnormal{ref}}$ produces completions with maximum likelihood for the preferred responses. This assures that we can effectively consider $\pi_{\textnormal{ref}} = \pi^{SFT}$.
    \item Similar to \acrshort{acr_rlhf}, initialize both the parameterized policy $\pi_{\theta}$ and reference policy $\pi_{\textnormal{ref}}$ by $\pi^{SFT}$.
    \item Optimize $\pi_{\theta}$ to minimize $\mathscr{L}_{DPO}$ with an appropriate value for $\beta$.
\end{enumerate}

\section{Methodology}
\paragraph{Codebase Note}

An interested reader can find the notebooks used for creating datasets and human preferences alignment described in the forthcoming sections at \href{https://bitbucket.org/paladinanalytics/notebooks}{https://bitbucket.org/paladinanalytics/notebooks}.

\subsection{Dataset Creation}
\label{methodology:dataset_creation}
For formulating instruction/response pairs to fine-tune Open-LLaMA 3B v2 \cite{openlm2023openllama}
for instruction following, we employed various dataset generation methods. These original research
methodologies depended on OpenAI's GPT models for their respective dataset generation. However, to
promote open source practices, we selected models without restrictions on commercial use of their
generated content. Through our exploration, we identified several suitable models and ultimately
chose \texttt{h2oai/h2ogpt-gm-oasst1-en-2048-falcon-40b-v2} \cite{h2oaifalcon40boasst}.

In both the LaMini and Evol-Instruct methods, we utilized the \texttt{databricks/databricks-dolly-15k} dataset \cite{DatabricksBlog2023DollyV2} to select seed
instructions as examples for the new dataset. This dataset contains instruction/response pairs
suitable for instruction-tuning pretrained models, dispersed among the following categories:
Creative Writing, Closed Question Answering (QA), Open QA, Summarization, Information Extraction,
Classification, and Brainstorming.

For the Orca scheme, we used the FLAN-v2 Collection \cite{longpre2023flan} to select query and
response pairs, following the methodology of the Orca paper authors \cite{mukherjee2023orca}. The
FLAN-v2 dataset comprises several submixtures, including Flan2021 (142 subtasks), T0 (193 subtasks),
Niv2 (1560 subtasks), CoT (18 subtasks), and Dialog. Each submixture contains multiple subtasks
covering a diverse range of NLP applications. As in the Orca paper, we sampled only zero-shot
queries for explanation generation and excluded the Dialog submixture.

\subsubsection{LaMini Dataset}
\label{subsubsec:lamini}
\paragraph*{Prompt for Instruction Generation}
To emulate the procedure established by \citeauthor{wu2023lamini} in \cite{wu2023lamini} for
composing their dataset, we initially used the same prompt they did with the \texttt{gpt-3.5-turbo} model.
However, we found that this prompt did not produce examples in the appropriate format with our
parent model. Therefore, we made slight alterations to devise an alternative fundamental prompt.
Figure \ref{fig:lamini_example_guided_prompt} displays an instance of an instruction generation
prompt that adheres strictly to an example-guided approach. To automate the dataset generation
process, the program randomly determined whether to use a topic-guided generation or not. If
selected, three arbitrary Wikipedia categories meeting the same criteria as in \cite{wu2023lamini}
were incorporated into the prompt. We provided a pre-set number of
three examples, randomly selected from the same instruction category, also decided randomly, within the dataset. This
instruction category was also incorporated into the prompt (refer to Figure
\ref{fig:lamini_example_guided_prompt}).

\begin{figure}[ht]
    \centering
    \fbox{
        \begin{minipage}{0.9\linewidth}
            \#\#\# SYSTEM: You are an AI assistant. Answer as honestly and correctly as possible.\\
            \#\#\# YOUR TASK: Generate 5 diverse examples that are similar to the provided examples.\\
            You do not need to provide responses to the generated examples.\\
            Do not repeat the provided examples.\\
            Each generated example must include an instruction.\\
            Each generated example may have an additional context if necessary.\\
            Each generated example can be either an imperative sentence or a question.\\
            Each generated example must begin with "<example>" and end with "</example>"\\

            \#\#\# PROVIDED EXAMPLES(Category: classification):\\
            <example>Identify which instrument is string or percussion: Cantaro, Gudok</example>\\
            <example>Classify each of the following as a primary color or a secondary color</example>\\
            <example>Which is a species of fish? Banjo or Guitar</example>\\

            \#\#\#RESPONSE:
        \end{minipage}
    }
    \caption{An example of an instruction generation prompt based on three random examples from \texttt{databricks-dolly-15k}}
    \label{fig:lamini_example_guided_prompt}
\end{figure}

\paragraph*{Prompt for Response Generation}
The approach used for generating responses was commensurate with the original methodology in
\cite{wu2023lamini}. As shown in Figure \ref{fig:lamini_response_prompt}, the generated instruction
was encapsulated within the prompt template before being processed by the model.

\begin{figure}[ht]
    \centering
    \fbox{
        \begin{minipage}{0.9\linewidth}
            \#\#\# SYSTEM: You are an AI chat assistant. Answer as honestly and correctly as possible. Do not use \#\#\# in your response.\\
            \#\#\# INSRUCTION: How does photosynthesis work and why is it important for plants and humans? Input:Photosynthesis is the process by which plants convert sunlight into energy. During photosynthesis, carbon dioxide from the air and water from the soil are converted into glucose, which provides food for the plant. Oxygen is released as a byproduct of this reaction. Photosynthesis is essential for plants because it provides them with the nutrients they need to grow and reproduce. It is also important for humans because it produces oxygen, which we need to breathe.\\

            \#\#\# RESPONSE:
        \end{minipage}
    }
    \caption{Response generation prompt used}
    \label{fig:lamini_response_prompt}
\end{figure}

\paragraph*{Instruction Generation}
In each iteration, the chosen model, \texttt{h2oai/h2ogpt-gm-oasst1-en-2048-falcon-40b-v2}, was
directed to construct five examples simultaneously. This number was intuitively chosen to avoid
surpassing the model's context limitation. After the examples were generated, a series of regular
expressions segmented the model response, creating a list containing the examples. This list, along
with a few related fields, was then saved into the dataset. Note that responses for the instructions
were not generated at this stage.

The above process was carried out iteratively, yielding a total of 1,504 instructions. This figure
does not hold any statistical relevance but was the maximum quantity we could effectively handle within
our resource constraints.

\paragraph*{Manual Inspection of the Instruction-Only Dataset}
Given the comparatively low parameter count of the utilized parent model, instances of inconsistencies in
model output across different iterations were reasonably anticipated \cite{naveed2023comprehensive}.
Consequently, we undertook the task of manually inspecting the generated dataset. Our efforts
revealed several issues:

\begin{enumerate}
    \item Although the model was guided to generate examples, each enclosed within "<example>" tags (Figure \ref{fig:lamini_example_guided_prompt}), approximately 53 examples in the
          dataset had all five examples enclosed within one set of markup tags, appearing as:
          \begin{itemize}
              \item "Here are 5 examples..."
              \item "Here are five examples..."
          \end{itemize}
          It's conceivable that the model encapsulated its entire response within the markup tags, resulting
          in the whole response being extracted during the regular expression matching process.
    \item Some instructions in our dataset were exactly equal to the seed instructions randomly chosen from the Dolly dataset.
\end{enumerate}

To rectify these discrepancies, we employed the following solutions:

\begin{enumerate}
    \item We used search queries to identify the instructions with the aforementioned starting phrases
          and manually extracted the individual examples as new entries.
    \item To identify equal examples, we used a two-step process:
          \begin{enumerate}
              \item We used the \texttt{SequenceMatcher} class from Python's \texttt{difflib} library
                    \cite{difflibPython}, which applies the Gestalt approach for pattern recognition
                    \cite{patternmatching:the_gestalt_approach_2013}, to find the instruction in the Dolly dataset most
                    similar to each entry in ours. The resulting similarity ratios were used in subsequent steps.
              \item We determined the Levenshtein distance \cite{Lev65} between the matched strings.
              \item After randomly scrutinizing examples, we derived that identical instances were those with a
                    similarity ratio of 0.6 or higher and a Levenshtein distance of 9 or less. These were removed from
                    the dataset.
          \end{enumerate}
\end{enumerate}

\paragraph*{Response Generation}
After generating all the instructions and manual inspection of the instructions generated, the model was instructed to generate responses pertinent to these
instructions using the prompt shown in Figure \ref{fig:lamini_response_prompt}.

\subsubsection{Evol-Instruct Dataset}
\label{subsubsec:evol_instruct}
\paragraph*{Prompt for Instruction Generation}
Figures \ref{fig:evol_in_depth_evolving_prompt_example} and
\ref{fig:evol_in_breadth_evolving_prompt_example} illustrate the designs of the prompts used in our
version of evol-instruct, tailored for our parent model. While employing the
same model as in LaMini (Section \ref*{subsubsec:lamini}), we designed the prompts to adhere to a
conversational syntax. This conversational style was chosen due to the fact that the prompting style used in our implementation of the LaMini scheme did not yield instructions of a comparable quality here.

\begin{figure}[ht]
    \centering
    \fbox{
        \begin{minipage}{0.9\linewidth}
            <human>: I want you to act as a prompt rewriter.\\
            Your objective is to rewrite the \#Given Prompt\# into a more complex version.\\
            But the rewritten prompt must be reasonable and must be understood and responded by humans.\\
            Your rewriting cannot omit the non-text parts such as the table and code in \#Given Prompt\#:. Also, please do not omit the context in \#Given Prompt\#.\\
            You should try your best not to make the \#Rewritten Prompt\# become verbose, \#Rewritten Prompt\# can only add 10 to 20 words into \#Given Prompt\#.\\
            '\#Given Prompt\#', '\#Rewritten Prompt\#', 'given prompt' and 'rewritten prompt' are not allowed to appear in \#Rewritten Prompt\#\\
            You SHOULD complicate the given prompt by adding one more constraints/requirements into \#Given Prompt\#\\
            \#Given Prompt\#:\\
            Why did Syd Barrett left the Pink Floyd?\\
            <bot>: \#Rewritten Prompt\#:
        \end{minipage}
    }
    \caption{An in-depth evolving prompt used to add constraints to a random instruction in \texttt{databricks-dolly-15k}}
    \label{fig:evol_in_depth_evolving_prompt_example}
\end{figure}

\begin{figure}[ht]
    \centering
    \fbox{
        \begin{minipage}{0.9\linewidth}
            <human>: I want you to act as a prompt creator.\\
            Your goal is to draw inspiration from the \#Given Prompt\# to create a brand new prompt.\\
            This new prompt should belong to the same domain as the \#Given Prompt\# but be even more rare.\\
            The LENGTH and difficulty level of the \#Created Prompt\# should be similar to that of the \#Given Prompt\#.\\
            The \#Created Prompt\# must be reasonable and must be understood and responded by humans.\\
            '\#Given Prompt\#', '\#Created Prompt\#', 'given prompt' and 'created prompt' are not allowed to appear in \#Created Prompt\#.\\
            Your response only contains the \#Created Prompt\# and no explanation of the new prompt. Do not provide a response to either the \#Given Prompt\# or the \#Created Prompt\#.\\
            \#Given Prompt\#:\\
            Which episodes of season four of Game of Thrones did Michelle MacLaren direct?\\
            <bot>: \#Created Prompt\#:
        \end{minipage}
    }
    \caption{An in-breadth evolving prompt based on a random instruction in \texttt{databricks-dolly-15k}}
    \label{fig:evol_in_breadth_evolving_prompt_example}
\end{figure}

\paragraph*{Prompt for Response Generation}
Consistent with the procedure outlined in the LaMini paradigm for response generation, we adopt an
analogous approach by inputting the generated instruction into the model using a conversational
format, as shown in Figure \ref*{fig:evol_response_prompt_example}.

\begin{figure}[ht]
    \centering
    \fbox{
        \begin{minipage}{0.9\linewidth}
            <human>: Investigate the relationship between childhood inquisitiveness and adult inquisitiveness by examining the ways in which children's questions can be transformed into curiosity about the world and how this curiosity can evolve throughout their lives. Provide examples of how parents, caregivers, and educators can nurture children's natural curiosity and encourage them to explore different topics. Discuss potential benefits and challenges that come with having an inquisitive mind as one grows older, including the development of critical thinking skills and the tendency to question authority.\\
            <bot>:
        \end{minipage}
    }
    \caption{The prompt template used for response generation in the evol-instruct dataset generation process}
    \label{fig:evol_response_prompt_example}
\end{figure}

\paragraph*{Prompt for Equality Check}
The equality check prompt was initially designed with a clear directive for the model to respond
with either 'equal' or 'not equal'. However, during execution, we observed instances where the
model's responses deviated from these expected binary choices. To address this issue, we engineered
the prompt to mimic the first few tokens of the model's response. Figure
\ref*{fig:evol_equality_check_prompt} illustrates this, providing the initial instruction and the
potential evolved prompt. In most cases, the model's response aligned with one of the requested
choices: 'equal' or 'not equal'. This alignment was crucial for the automated script designed to
identify the evolution of the instruction based on these specific responses.

\begin{figure}[ht]
    \centering
    \fbox{
        \begin{minipage}{0.9\linewidth}
            <human>: Do you think the following two instructions are equal to each other in that they meet the following requirements:\\
            1. They have same constraints and requirements.\\
            2. They have same depth and breadth of the inquiry.\\
            The First Prompt: How did Andy Warhol create the "piss paintings"?\\
            The Second Prompt: What are some of the techniques employed by Andy Warhol in creating his famous "piss paintings", and what was the significance of these works in the history of art?\\
            Your response should be either equal or not equal.\\
            <bot>: The two prompts are
        \end{minipage}
    }
    \caption{Equality check prompt used in the evol-instruct scheme}
    \label{fig:evol_equality_check_prompt}
\end{figure}

\paragraph*{Categorical Subsets of Evolution}
The initial Dolly dataset comprises approximately 15,000 instructions. Our evolution process
involved selecting a subset of 100 instructions from a single category and subjecting them to
evolution for a maximum of two epochs, with the number of epochs determined randomly for each
category. This process was applied to all categories, with each category undergoing multiple
iterations.

Notably, the evolution strategy (in-depth or in-breadth) was chosen randomly by the Python script
for each iteration. The selection of the specific in-depth evolving operation followed a similar
random process.

\begin{figure}[ht]
    \centering
    \includegraphics[width=0.6\linewidth]{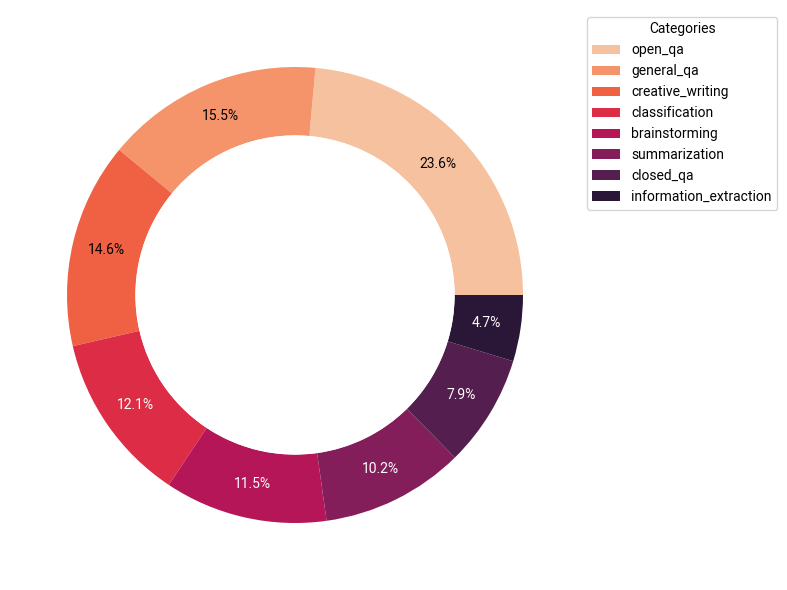}
    \caption{Distribution of categories in the evol-instruct dataset}
    \label{fig:evol_category_distribution}
\end{figure}

\paragraph*{Distribution of Categories}
Unlike the approach in the LaMini paradigm, we maintained a record of the category for each evolved
instruction. As depicted in Figure \ref*{fig:evol_category_distribution}, the distribution of
categories is nearly uniform, with \texttt{open\_qa} representing a higher outlier and
\texttt{information\_extraction} a lower one.

Closer examination reveals that instructions associated with an input, such as
\texttt{information\_extraction}, \texttt{closed\_qa}, and \texttt{summarization}, are more sparsely
distributed. This can be directly attributed to the model's limitations in handling extensive
context sizes and accurately following instructions, leading to the non-evolution of a majority of
examples with an additional context.

\begin{figure}[ht]
    \centering
    \begin{minipage}{.48\textwidth}
        \centering
        \includegraphics[width=\linewidth]{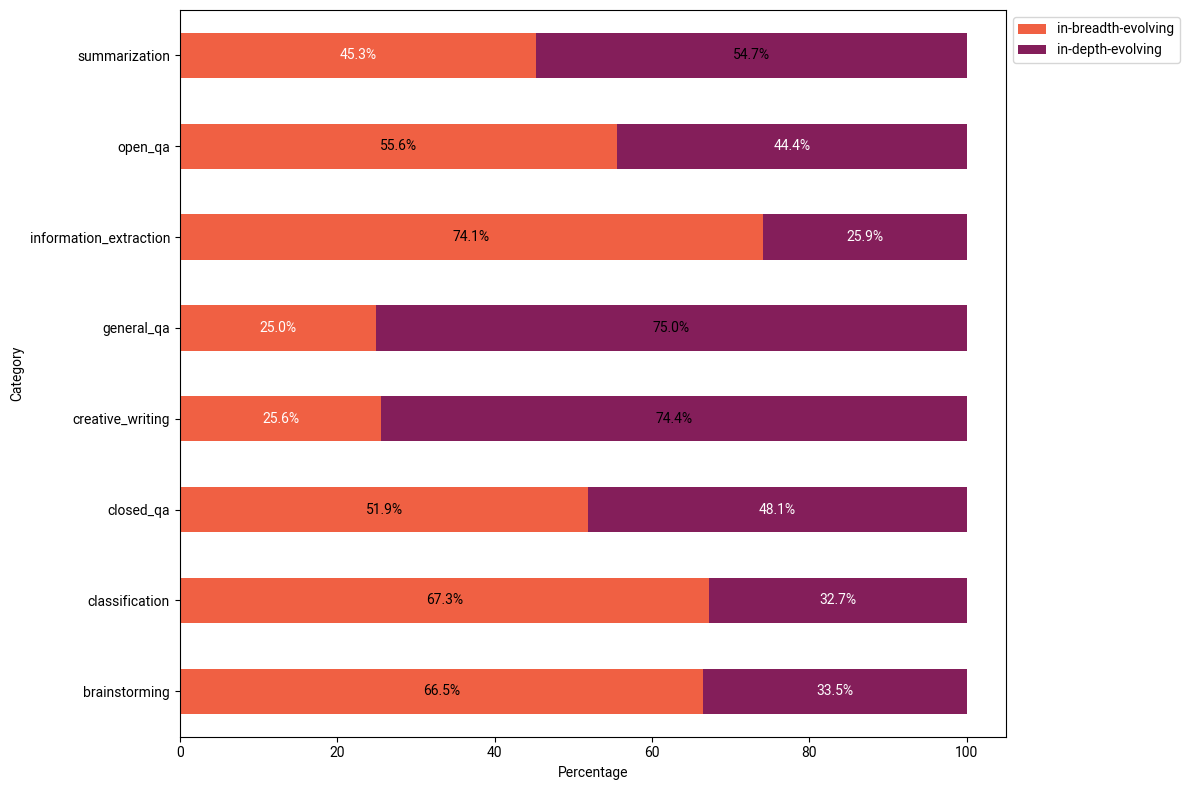}
        \caption{Distribution of evolution strategy of each category}
        \label{fig:evol_evolution_strategy}
    \end{minipage}
    \hfill
    \begin{minipage}{.48\textwidth}
        \centering
        \includegraphics[width=\linewidth]{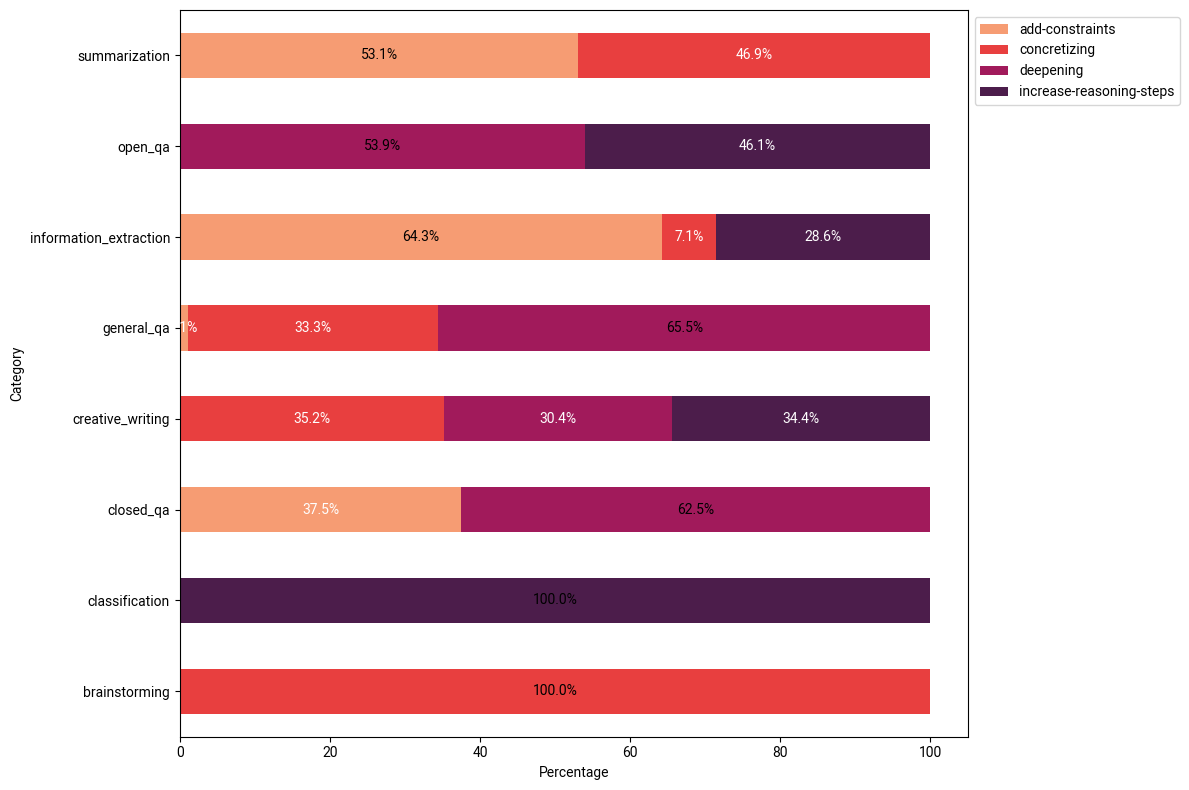}
        \caption{Distribution of evolution operation of each category for in-depth-evolving strategy}
        \label{fig:in-depth-evolving}
    \end{minipage}
\end{figure}

Figure \ref*{fig:evol_evolution_strategy} illustrates the distribution of evolution strategies
across each category. Despite the random binary choice of evolution strategy, which would suggest a
roughly equal distribution, the stacked bar chart reveals a different scenario. This discrepancy can
be attributed to the varying number of epochs for which different subsets underwent evolution.

Figure \ref*{fig:in-depth-evolving} shows the distribution of evolution operations executed under
the in-depth-evolving strategy. The uneven distribution aligns with the practice of applying the
same in-depth-evolving operation to all instructions within a single evolution epoch. As not all
subsets undergo the same number of evolution epochs, the resulting dataset exhibits this
particular distribution.

\subsubsection{ORCA Dataset}
\label{subsubsec:orca}
Given the time and compute constraints we sampled maximum 3 response - query pairs for each subtask in 2 of the 4 FLAN submixtures. Namely T0 and Niv2. In cases where there were less than 3 response-query pairs we sampled the maximum available. For Flan2021 and T0 submixture we adhered to the same sampling algorithm as the authors of Orca.

\begin{figure}[ht]
    \centering
    \begin{tabular}{l}
        \hline
        \textbf{Algorithm 1:} Sampling Algorithm for Flan 2021 and T0 collection.        \\
        \hline
        \textbf{Input:} tasks $T = \{t_1,t_2,...,t_m\}$, number of queries to sample $n$ \\
        \textbf{Output:} sampled queries $Q = \{q_1,q_2,...,q_n\}$                       \\
        $Q \gets$ empty list                                                             \\
        \textbf{while} $|Q| < n$ \textbf{do}                                             \\
        \quad $t \gets$ randomly sample a task from $T$                                  \\
        \quad $q \gets$ randomly sample a query without replacement from $t$             \\
        \quad add $q$ to $Q$                                                             \\
        \quad \textbf{if} $t$ is empty \textbf{then}                                     \\
        \quad \quad remove $t$ from $T$                                                  \\
        \quad \textbf{end}                                                               \\
        \textbf{end}                                                                     \\
        \textbf{return} $Q$                                                              \\
        \hline
    \end{tabular}
    \caption{Sampling Algorithm for Flan 2021 and T0 collection adapted from \cite{mukherjee2023orca}.}
\end{figure}

We employed the same system messages that were used in the orca \cite{mukherjee2023orca} to elicit detailed and explained responses to the queries from the FLAN collection via the model we selected, \texttt{h2oai/h2ogpt-gm-oasst1-en-2048-falcon-40b-v2}, with the exception of system message id 1: <empty system message>.

\begin{table}[ht]
    \centering
    \begin{tabular}{p{0.05\textwidth}p{0.80\textwidth}}
        \toprule
        \textbf{Id.} & \textbf{System Message}                                                                                                                                                                                                                                                                                                                                                                                                         \\
        \midrule
        1            & <empty system message>                                                                                                                                                                                                                                                                                                                                                                                                          \\
        2            & You are an AI assistant. Provide a detailed answer so user don't need to search outside to understand the answer.                                                                                                                                                                                                                                                                                                               \\
        3            & You are an AI assistant. You will be given a task. You must generate a detailed and long answer.                                                                                                                                                                                                                                                                                                                                \\
        4            & You are a helpful assistant, who always provide explanation. Think like you are answering to a five year old.                                                                                                                                                                                                                                                                                                                   \\
        5            & You are an AI assistant that follows instruction extremely well. Help as much as you can.                                                                                                                                                                                                                                                                                                                                       \\
        6            & You are an AI assistant that helps people find information. Provide a detailed answer so user don't need to search outside to understand the answer.                                                                                                                                                                                                                                                                            \\
        7            & You are an AI assistant. User will you give you a task. Your goal is to complete the task as faithfully as you can. While performing the task think step-by-step and justify your steps.                                                                                                                                                                                                                                        \\
        8            & You should describe the task and explain your answer. While answering a multiple choice question, first output the correct answer(s). Then explain why other answers are wrong. Think like you are answering to a five year old.                                                                                                                                                                                                \\
        9            & Explain how you used the definition to come up with the answer.                                                                                                                                                                                                                                                                                                                                                                 \\
        10           & You are an AI assistant. You should describe the task and explain your answer. While answering a multiple choice question, first output the correct answer(s). Then explain why other answers are wrong. You might need to use additional knowledge to answer the question.                                                                                                                                                     \\
        11           & You are an AI assistant that helps people find information. User will you give you a question. Your task is to answer as faithfully as you can. While answering think step-by step and justify your answer.                                                                                                                                                                                                                     \\
        12           & User will you give you a task with some instruction. Your job is follow the instructions as faithfully as you can. While answering think step-by-step and justify your answer.                                                                                                                                                                                                                                                  \\
        13           & You are a teacher. Given a task, you explain in simple steps what the task is asking, any guidelines it provides and how to use those guidelines to find the answer.                                                                                                                                                                                                                                                            \\
        14           & You are an AI assistant, who knows every language and how to translate one language to another. Given a task, you explain in simple steps what the task is asking, any guidelines that it provides. You solve the task and show how you used the guidelines to solve the task.                                                                                                                                                  \\
        15           & Given a definition of a task and a sample input, break the definition into small parts. Each of those parts will have some instruction. Explain their meaning by showing an example that meets the criteria in the instruction. Use the following format:\newline Part \#: a key part of the definition.\newline Usage: Sample response that meets the criteria from the key part. Explain why you think it meets the criteria. \\
        16           & You are an AI assistant that helps people find information.                                                                                                                                                                                                                                                                                                                                                                     \\
        \bottomrule
    \end{tabular}
    \caption{System messages used by the authors of orca to elicit detailed responses from the LFM}
    \label{table:orca-system-messages}
\end{table}

Given that certain system messages are better suited on certain sub mixtures we chose the sampling distribution showcased in Figure \ref*{table:orca-sampling-distribution} when randomly sampling a system message for a query.

\begin{table}[ht]
    \centering
    \begin{tabular}{p{0.3\textwidth}p{0.3\textwidth}p{0.3\textwidth}}
        \toprule
        \textbf{Submixture} & \textbf{Message Id}               & \textbf{Probability}                                                                            \\
        \midrule
        COT                 & 6, 11, 16                         & Uniform probability of $\nicefrac{1}{3}$ each                                                   \\
        NiV2                & 1, 2, 5, 7, 9, 12, 13, 14, 15, 16 & Uniform probability of $\nicefrac{1}{9}$ each                                                   \\
        T0                  & 1, 2, 3, 5, 7                     & Uniform probability of $\nicefrac{1}{5}$ each                                                   \\
        FLAN2021            & 3, 4, 7, 8, 9                     & [$\nicefrac{1}{4}$, $\nicefrac{1}{4}$, $\nicefrac{1}{4}$, $\nicefrac{1}{4}$, $\nicefrac{1}{4}$] \\
        \bottomrule
    \end{tabular}
    \caption{System messages suited for each submixture and its sampling probability}
    \label{table:orca-sampling-distribution}
\end{table}

Although the original method for prompting the LFM in orca followed a system-instruction-response format, we opted to create our own prompt in hopes of better aligning the LFMs output with the expected output by following a system-instruction-expected output-response format.

\begin{figure}[ht]
    \centering
    \fbox{
        \begin{minipage}{0.9\linewidth}
            \#\#\# \{system\_msg\} \\
            \#\#\# your task is:\\
            \{query\}
            \\
            \#\#\# the correct answer to this task is:\\
            \{target\}\\
            use this correct answer to guide you.
            \\
            \#Response:
        \end{minipage}
    }
    \caption{Prompt used during Orca}
    \label{fig:orca-prompt}
\end{figure}

An example of such a prompt used during this process is shown in Figure \ref*{fig:orca-prompt-example}.

\begin{figure}[ht]
    \centering
    \fbox{
        \begin{minipage}{0.9\linewidth}
            \#\#\# You are an AI assistant that helps people find information. User will you give you a question. Your task is to answer as faithfully as you can. While answering think step-by-step and justify your answer.
            \\
            \#\#\# your task is:\\
            Of the following two sentences, which one is against common sense? Options: - Sentence A: "He poured orange juice on his cereal." - Sentence B: "He poured milk on his cereal." Let's reason step by step:
            \\
            \#\#\# the correct answer to this task is:\\
            Orange juice does not taste good on cereal. Final answer: Sentence A.
            use this correct answer to guide you.
            \\
            \#Response:
        \end{minipage}
    }
    \caption{Example prompt used during Orca}
    \label{fig:orca-prompt-example}
\end{figure}

Following the stated method we generated a total of 5507 explanation tuning data samples adhering to the orca scheme. The final count distribution of detailed orca scheme responses for each submixture are shown in Figure \ref*{table:orca-response-distribution}.

\begin{table}[ht]
    \centering
    \begin{tabular}{p{0.45\textwidth}p{0.45\textwidth}}
        \toprule
        \textbf{Submixture} & \textbf{Response duration} \\
        \midrule
        T0                  & 579                        \\
        COT                 & 54                         \\
        Flan 2021           & 210                        \\
        Niv 2               & 4665                       \\
        \midrule
        \textbf{Total}      & \textbf{5507}              \\
        \bottomrule                                      \\
    \end{tabular}
    \caption{Distribution of detailed responses for each submixture intended to be used for explanation tuning following the orca methodology}
    \label{table:orca-response-distribution}
\end{table}

\subsection{Rejection Sampling}
\label{methodology:rejection_sampling}
Following the execution of each dataset generation scheme described in Section \ref{methodology:dataset_creation}, we conducted a rejection sampling process to eliminate instances that did not fit a specified criteria. During this process, we utilized GPT-4 as a judge to assess the quality and appropriateness of the instructions. The system and human prompts employed in the rejection phase are depicted in Figures \ref{fig:rejection_system_prompts} and \ref{fig:rejection_human_prompts}, respectively. These figures illustrate the specific prompts used to guide GPT-4's evaluation of the generated instructions. The criteria for rejection can be seen in bold font. It is important to note that the two different prompt formats are a result of the Orca dataset's structural differences compared to the other two datasets, which necessitated a distinct prompt template to the rejection sampling process for Orca.

\begin{figure}[ht]
    \centering
    \fbox{
        \begin{minipage}{0.9\linewidth}
            I want you to act as an expert instruction/response evaluator.\\
            You are given an instruction and a response below.\\
            The instruction is within <instruction> and </instruction> tags, and the response is within <response> and </response> tags.\\
            \textbf{Your task is to evaluate whether the given response contains sufficient information to be clear, complete and specific to the given instruction.}\\
            You should also rate the response on a scale of 1 to 7, 1 being the worst and 7 being the best.\\
            If it is suitable, you should output <status>Accept</status>, rating within <rating> and </rating> and a reasoning for this status, rating within <reason> and </reason>.\\
            If it is not suitable, you should output <status>Reject</status> rating within <rating> and </rating> and a reasoning for this status, rating within <reason> and </reason>.\\
            Your response should contain none other than the status, rating and reason.\\
        \end{minipage}
    }
    \fbox{
        \begin{minipage}{0.9\linewidth}
            I want you to act as an expert prompt/response evaluator.\\
            You are given an instruction and a corresponding expected response. You are also given the generated response from an LLM for the same instruction.\\
            The instruction is within <instruction> and </instruction> tags, the expected response is within <expected> and </expected> tags, and the generated response is within <generated> and </generated> tags.\\
            \textbf{Your task is to evaluate whether the generated response is an accurate explanation of the expected response for the given instruction.}\\
            You should also rate the generated response on a scale of 1 to 7, 1 being the worst and 7 being the best.\\
            If it is an accurate explanation, the status of the response should be "Accept", and "Reject", if not.\\
            Your response should be in the following format:\\
            <status>Accept/Reject</status>\\
            <ratingInteger Rating between 1 and 7</rating>\\
            <reason>Your reasoning for status and rating</reason>\\
        \end{minipage}
    }
    \caption{System prompts used with GPT-4 for the evaluation phase. The one on the top depicts the system prompt used for LaMini/Evol-Instruct and the one on the bottom is for Orca. The bold font depicts the specific criteria for the rejection of dataset instances.}
    \label{fig:rejection_system_prompts}
\end{figure}

\begin{figure}[ht]
    \centering
    \fbox{
        \begin{minipage}{0.9\linewidth}
            <instruction>\{instruction\}</instruction>\\
            <response>\{response\}</response>
        \end{minipage}
    }
    \fbox{
        \begin{minipage}{0.9\linewidth}
            <instruction>\{inputs\}</instruction>\\
            <expected>\{targets\}</expected>\\
            <generated>\{explained\_targets\}</generated>
        \end{minipage}
    }
    \caption{Human prompt used with GPT-4 for the rejection sampling phase. The order is same as that in Figure \ref*{fig:rejection_system_prompts}}
    \label{fig:rejection_human_prompts}
\end{figure}

We tabulate the corresponding outcome of the rejection sampling phase in Table \ref{table:rejection-sampling-results}. We also decided to include the percentage of examples that yielded a blank response from GPT-4, which is shown in the last column. These results show that roughly $\nicefrac{3}{4}$ of the LaMini and Evol-Instruct datasets were accepted while only a quarter of the Orca dataset was accepted. It is worthwhile noting that almost half of this dataset were left undecided. Manual inspection of these examples suggest that most of them were too long and often contained gibberish.

\begin{table}[ht]
    \centering
    \begin{tabular}{lccc}
        \toprule
        \textbf{Dataset} & \multicolumn{1}{l}{\textbf{\# of accepted examples}} & \multicolumn{1}{l}{\textbf{\% of accepted examples}} & \multicolumn{1}{l}{\textbf{\% of  examples left undecided}} \\
        \midrule
        LaMini           & 1120                                                 & 74.5                                                 & 0.1                                                         \\
        Evol-Instruct    & 1567                                                 & 68.0                                                 & 0                                                           \\
        Orca             & 921                                                  & 16.7                                                 & 47.5                                                        \\
        \bottomrule                                                                                                                                                                                  \\
    \end{tabular}
    \caption{Number of accepted examples through the rejection sampling phase of each dataset and their respective percentages. The last column shows the percentage of examples that yielded a blank response from GPT.}
    \label{table:rejection-sampling-results}
\end{table}

\subsection{Finetuning}
\label{methodology:finetuning}
In our three-phase finetuning strategy, we sequentially finetuned the base model with the datasets described in Section \ref*{methodology:dataset_creation}, ordering them in the following arbitrary order: LaMini, Orca, and Evol-Instruct.

The three datasets shared a similar structure, consisting primarily of instruction/response pairs. However, the Orca dataset occasionally included system prompts. To maintain consistency during fine-tuning, we adapted a standard prompt template, as illustrated in Figure \ref*{fig:finetuning-prompt-template}. For the LaMini and Evol-Instruct datasets, which lacked system prompts, we employed a modified version of the Alpaca system prompt, shown in Figure \ref*{fig:default-system-prompt}.

\begin{figure}[H]
    \centering
    \fbox{
        \begin{minipage}{0.9\linewidth}
            \#\#\# System:
            \{system\}\\

            \#\#\# Instruction:
            \{instruction\}\\

            \#\#\# Response:
        \end{minipage}
    }
    \caption{The general prompt template adapated to fit all three finetuning schemes}
    \label{fig:finetuning-prompt-template}
\end{figure}

\begin{figure}[H]
    \centering
    \fbox{
        \begin{minipage}{0.9\linewidth}
            Below is an instruction that describes a task, optionally paired with an input that provides further context following that instruction. Write a response that appropriately completes the request.
        \end{minipage}
    }
    \caption{Default system prompt used in LaMini and evol-instruct}
    \label{fig:default-system-prompt}
\end{figure}

\paragraph*{Experimental Setup}

Our main focus during the finetuning phase was to efficiently fine-tune the base model, specifically for resource-constrained training environments, in contrast to the fully supervised fine-tuning approach employed by several major instruction models \cite{jiang2023mistral,touvron2023llama,penedo2023refinedweb}. To this end, we focused on utilizing the Q-LoRA fine-tuning algorithm \cite{dettmers2023qlora}.

\begin{figure}[ht]
    \centering
    \begin{subfigure}{0.45\textwidth}
        \includegraphics[width=\linewidth]{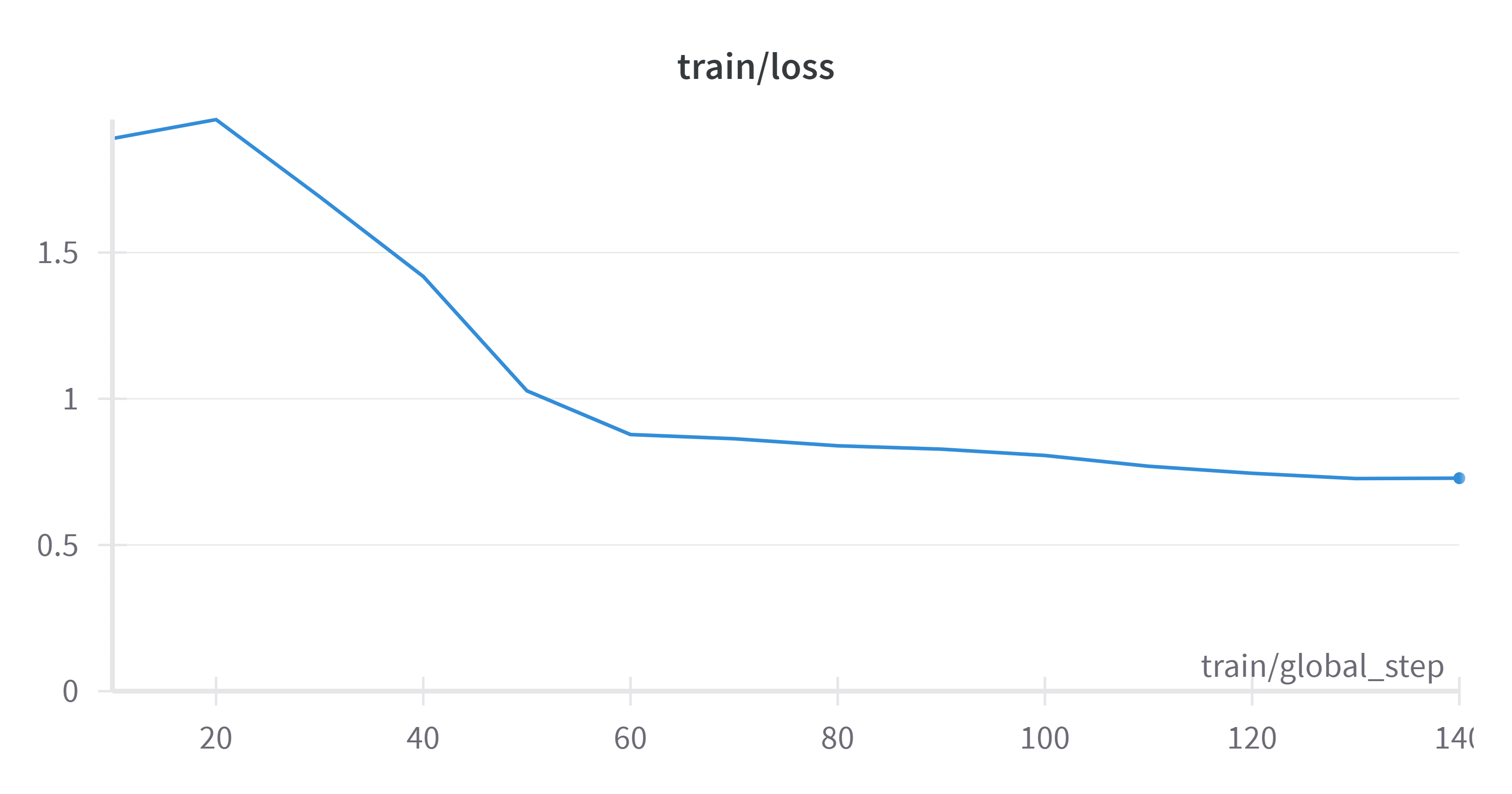}
        \caption{Train Loss during Run \#1}
    \end{subfigure}
    \begin{subfigure}{0.45\textwidth}
        \includegraphics[width=\linewidth]{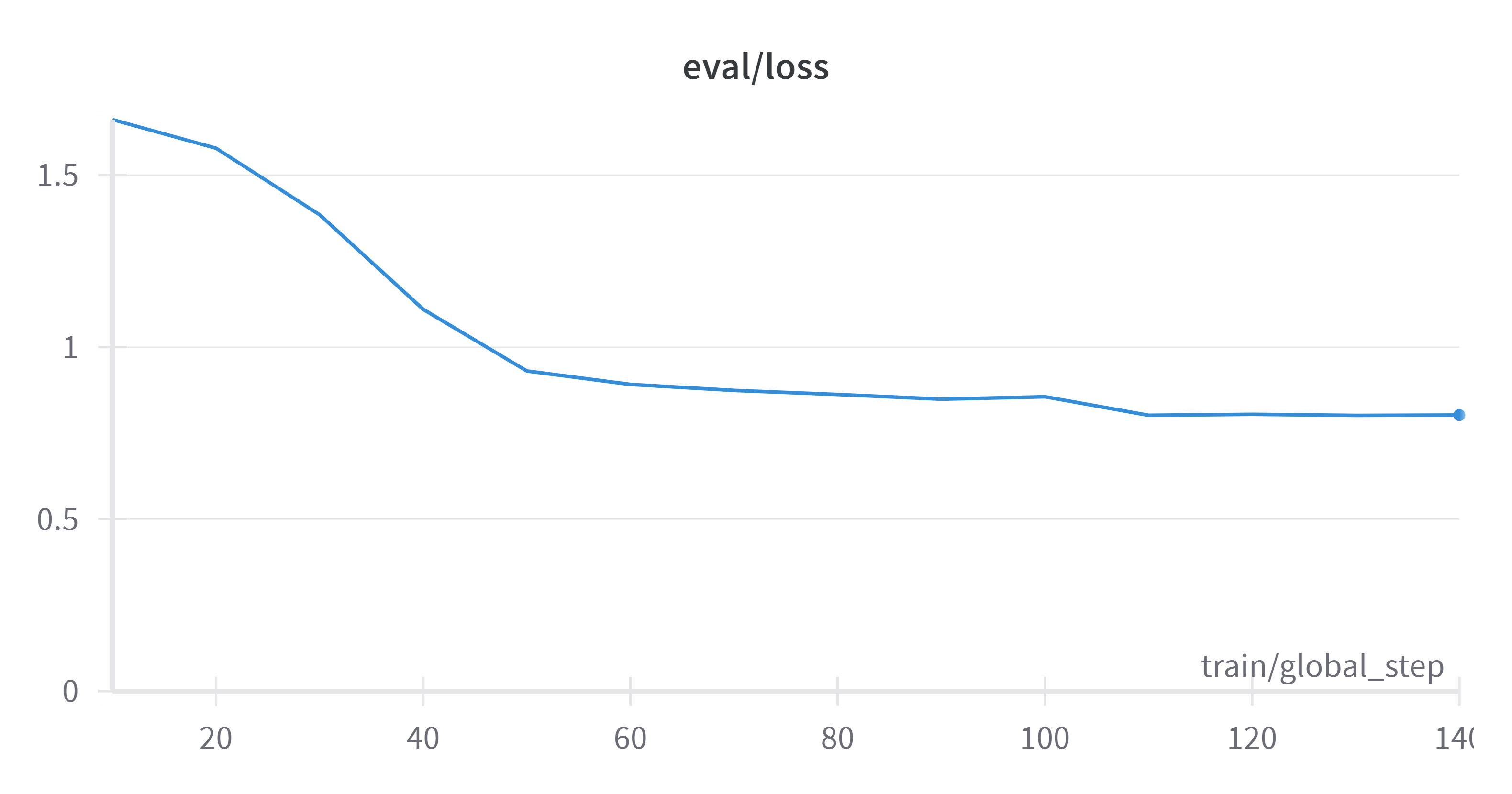}
        \caption{Eval Loss during Run \#1}
    \end{subfigure}

    % Add some space between the two rows of images
    \vspace*{2em}

    \begin{subfigure}{0.45\textwidth}
        \includegraphics[width=\linewidth]{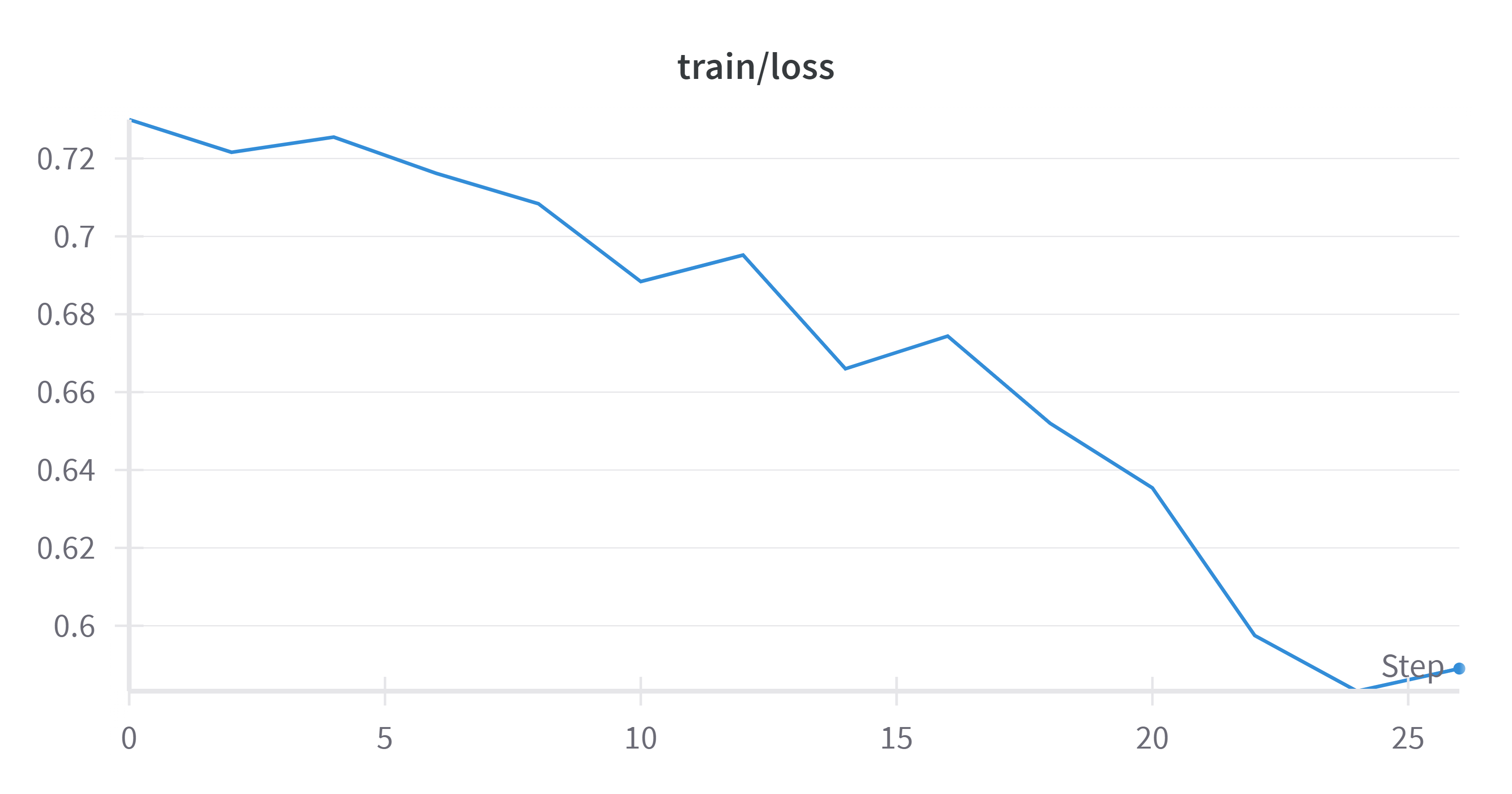}
        \caption{Train Loss during Run \#2}
    \end{subfigure}
    \begin{subfigure}{0.45\textwidth}
        \includegraphics[width=\linewidth]{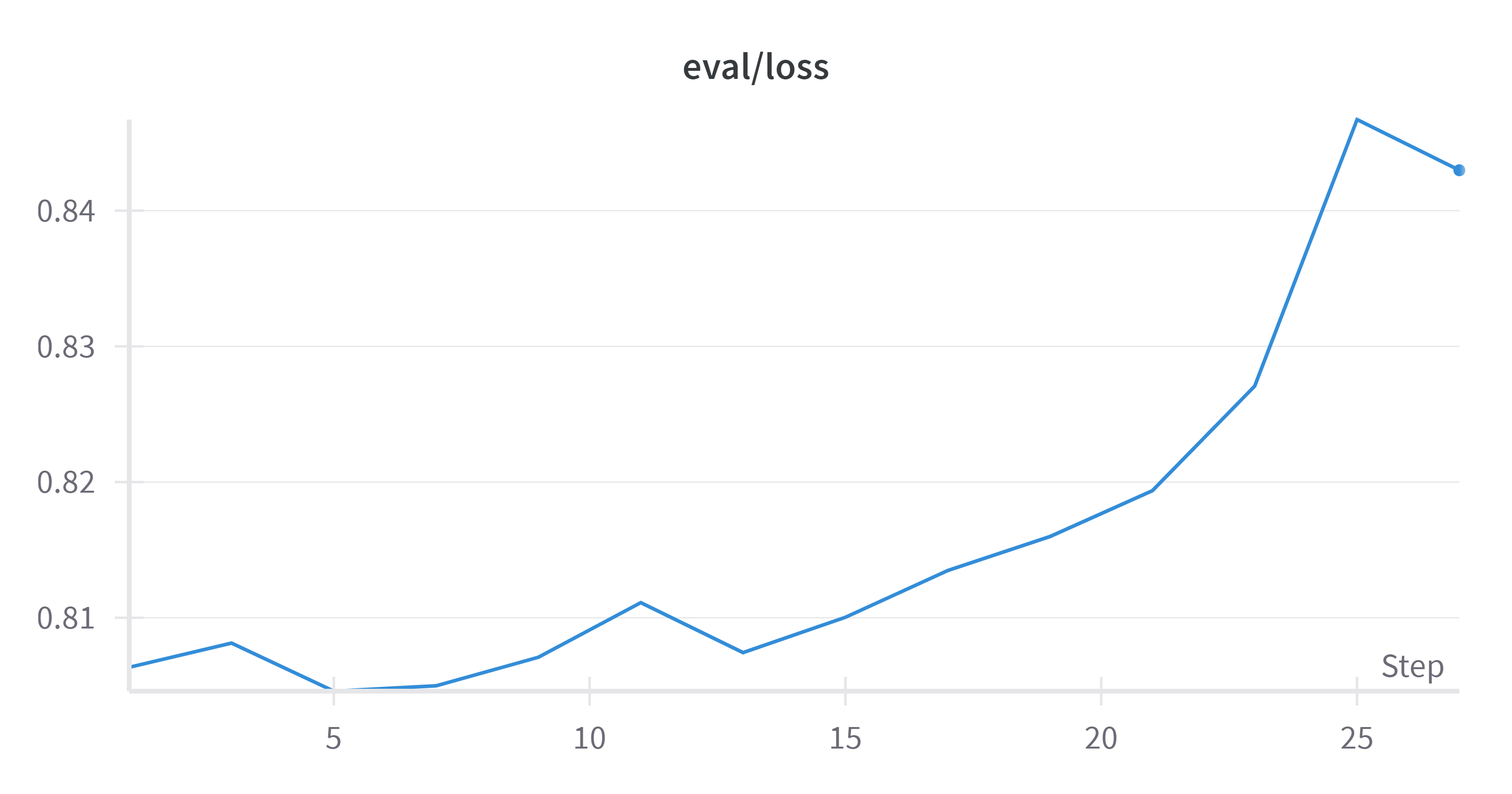}
        \caption{Eval Loss during Run \#2}
    \end{subfigure}

    \caption{Train and Eval loss during LaMini finetuning}
    \label{fig:lamini-loss-plots}
\end{figure}

For each dataset, we conducted fine-tuning in epochs of 10 until a clear divergence between training and evaluation loss was observed. As shown in Table \ref{table:finetuning-setup}, this approach led to a maximum of two runs per fine-tuning scheme. Figure \ref{fig:lamini-loss-plots} illustrates that during the LaMini fine-tuning, the second run resulted in a significant separation between training and evaluation losses. In contrast, during the second Orca fine-tuning phase, the evaluation loss initially followed a trajectory similar to the training loss for a few epochs before diverging. We opted out of a second run during the Evol-Instruct fine-tuning due to the losses diverging within the first run itself. Refer to Appendix \ref*{appendix:q_lora_finetuning_charts} for loss charts corresponding to the latter two fine-tuning scheme.

\begin{table}[H]
    \centering
    \begin{tabular}{@{}cccccc@{}}
        \toprule
        \textbf{Scheme} & \textbf{Run \#} & \textbf{\# of Epochs} & \textbf{Batch Size} & \textbf{Starting eval\_loss} & \textbf{Final eval\_loss} \\ \midrule
        LaMini          & 1               & 10                    & 64                  & 1.661                        & 0.8023                    \\
        LaMini          & 2               & 10                    & 64                  & 0.8063                       & 0.843                     \\ \addlinespace
        Orca            & 1               & 10                    & 64                  & 2.315                        & 1.539                     \\
        Orca            & 2               & 10                    & 64                  & 1.553                        & 1.533                     \\ \addlinespace
        Evol            & 1               & 10                    & 64                  & 1.006                        & 0.8835                    \\ \bottomrule
    \end{tabular}
    \caption{Finetuning setup followed in LaMini, Orca, and evol-instruct.}
    \label{table:finetuning-setup}
\end{table}

\paragraph*{Codebase}

We selected the code repository referenced in \citeurl{Vihangd} \cite{Vihangd}, which implements the Q-LoRA algorithm and supports fine-tuning various LLMs, including our chosen base model, Open-LLaMA 3B v2. We adapted this repository to better align with our requirements by modifying it to accommodate our dataset structure and fine-tuning prompt template. The customized repository has been made publicly available at \url{https://bitbucket.org/paladinanalytics/qlora-finetuning}.

\subsection{Human Preferences Alignment}
\label{methodology:human_pref_align}
We relied on \acrshort{acr_qlora} during every step of the \acrshort{acr_sft} recipe described above. Hence, every checkpoint is an adapter that is plugged in to the model when required. However, \acrshort{acr_dpo} update has been derived by considering the parameterization $\pi_{\theta}$ of the entire model\cite{rafailov2023direct}. It may very well be true that the approximation with low-rank adapters performs in a similar manner as it does in \acrshort{acr_sft}\cite{hu2021lora}, but it may also need further research. However, we considered the full parameterization approach by naively merging the adapter obtained at the end of the \acrshort{acr_sft} stage with the base model. The merged model after \acrshort{acr_sft} stage is denoted by $\pi_{m}$.\par

In our work, we chose to reuse Anthropic's HH-RLHF dataset\cite{bai2022training} for alignment fine-tuning. This is primarily based on the cost associated with the process of response pair generation and human preference labelling. The entire dataset contains nearly 161000 examples. Considering the resource constraints, we utilized the first 100000 examples and further split in 5:4 ratio between training (denoted by $\mathscr{D}_{\textnormal{HH}}$) and testing subsets respectively. As, $\mathscr{D}_{\textnormal{HH}}$ was not used during the \acrshort{acr_sft} stages, we initially fine tune $\pi_{m}$ with the preferred responses in $\mathscr{D}_{\textnormal{HH}}$ to mitigate the distribution shift. Since we do not use \acrshort{acr_qlora} in this stage to comply with the original work, none of the parameters were frozen. We perform \acrshort{acr_sft} for only 1 epoch as our previous experiments suggested that it was sufficient to account for the distribution shift. The resulting checkpoint is denoted by $\pi^{SFT}$. It is worthwhile to note that $\pi^{SFT}$ is the same as OpenBezoar-HH-RLHF-SFT.\par

Initializing both parameterized policy $\pi_{\theta}$ and the reference policy $\pi_{\textnormal{ref}}$ by $\pi^{SFT}$, we then performed \acrshort{acr_dpo} for 1 epoch over the same preference pairs in $\mathscr{D}_{\textnormal{HH}}$. We used $\beta = 0.1$ as guided by the default value in the experiments of \cite{rafailov2023direct}.\par

Additionally, after several attempts of trade-offs against the batch size (and gradient accumulation steps), we truncated/padded every example prompt to a length of 1024 and restricted the output maximum length to 512 in both stages of \acrshort{acr_dpo}.\par

\paragraph{Experimental Setup} We initially modified\footnote{Modified Implementation of \acrshort{acr_dpo}: \href{https://bitbucket.org/paladinanalytics/direct-preference-optimization}{https://bitbucket.org/paladinanalytics/direct-preference-optimization}} the official \acrshort{acr_dpo} implementation\footnote{Official Implementation of \acrshort{acr_dpo}: \href{https://github.com/eric-mitchell/direct-preference-optimization}{https://github.com/eric-mitchell/direct-preference-optimization}} to incorporate a few additional functionalities. Utilizing the relevant notebooks in our repository, we used DataCrunch\footnote{DataCrunch Cloud: \href{https://datacrunch.io/}{https://datacrunch.io/}} A100 1x80GB instances to conduct several test runs. However, owing to budget constraints and allocations to other experiments, fine-tuning was exclusively performed on Kaggle's 2xT4 runtimes. In order to cope with the limited run-time in Kaggle, we had to reduce the number of epochs to 1, as specified above. We employed the free-tier subscription of Weights \& Biases\footnote{Home Page: \href{https://wandb.ai/}{https://wandb.ai/}} for logging. Both implementations of \acrshort{acr_dpo} log the training loss for each batch. Hence, while the graph may exhibit fluctuations, an overall decreasing trend should be observed.\par

\begin{figure}[ht]
    \centering
    \includegraphics[width=0.5\textwidth]{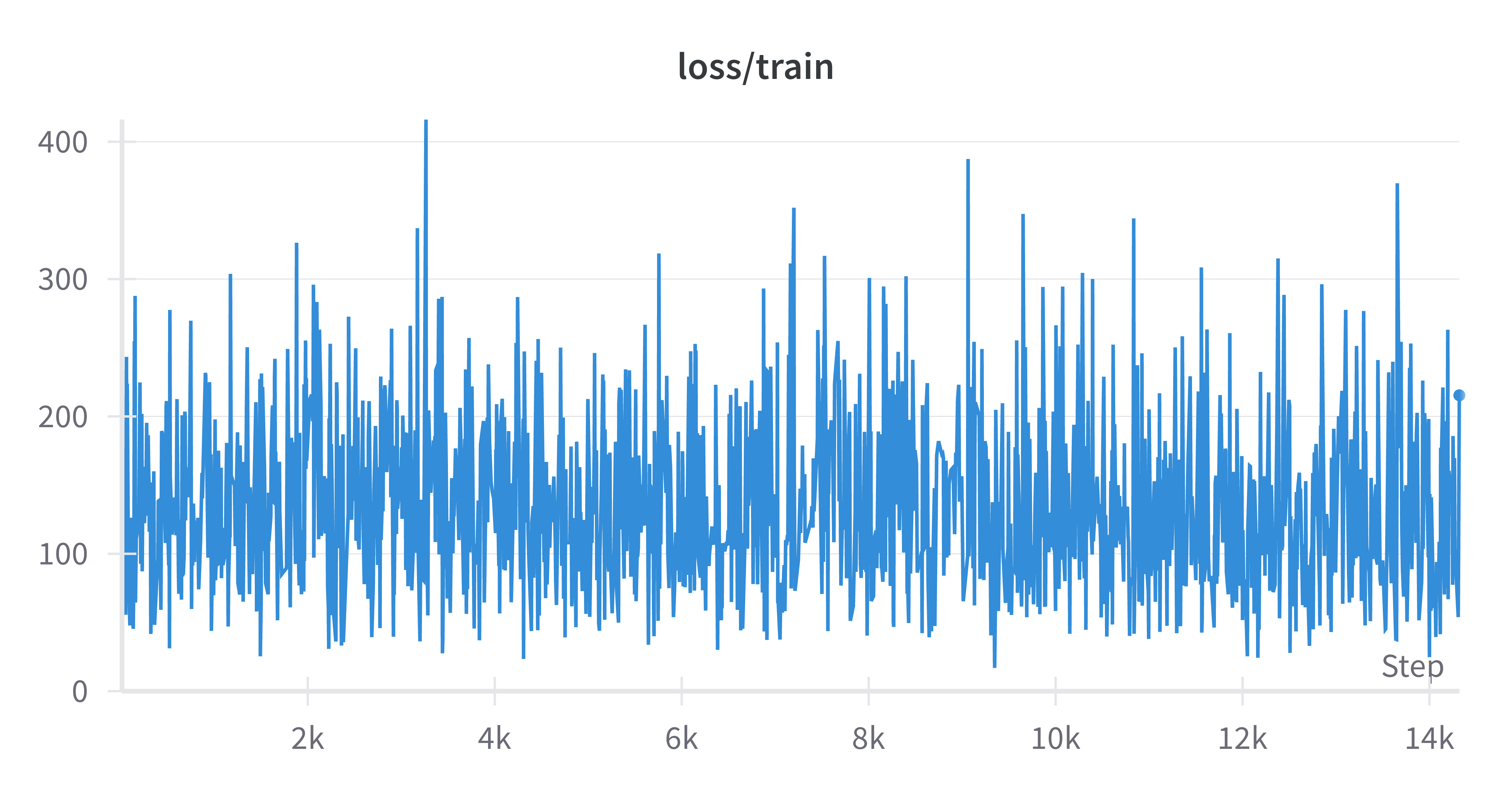}
    \caption{Training Loss for \acrshort{acr_dpo} on OpenBezoar-HH-RLHF-SFT. $x$-axis/``steps'' denotes the number of batches while $y$-axis denotes the average \acrshort{acr_dpo} loss}
    \label{fig:dpo_train_loss_train}
\end{figure}

\begin{figure}[ht]
    \centering
    \includegraphics[width=0.5\textwidth]{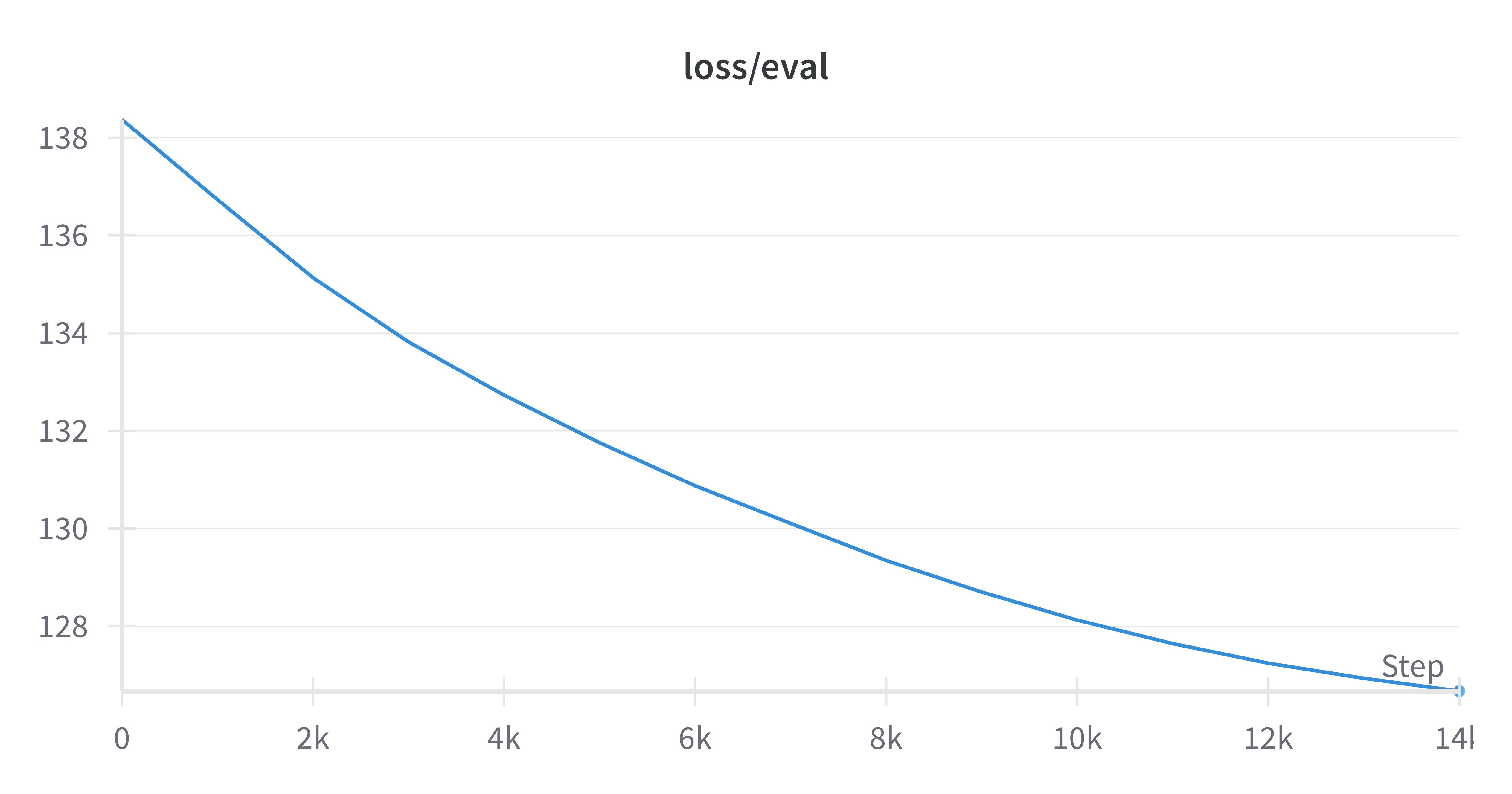}
    \caption{Evaluation Loss for \acrshort{acr_dpo} on OpenBezoar-HH-RLHF-SFT. $x$-axis/``steps'' denotes the number of batches while $y$-axis denotes the average \acrshort{acr_dpo} loss}
    \label{fig:dpo_train_loss_eval}
\end{figure}

\paragraph{Training Results} The training process lasted approximately 12 hours, with the checkpoints being saved in Kaggle's working environment. Figure \ref{fig:dpo_train_loss_train} illustrates the training loss, while Figure \ref{fig:dpo_train_loss_eval} depicts the evaluation loss. While the decreasing trend in the training loss may not be readily apparent, the decrease in the evaluation loss is observed exactly as expected. No degeneracy was observed with the premeditated value for $\beta$.\par

\paragraph{Future Work} The training process should be continued beyond just 1 epoch. Our goal was to simply evaluate the improvement over OpenBezoar-HH-RLHF-SFT due to the constraints we faced. Furthermore, a curious reader may also choose to investigate the use of \acrshort{acr_lora} for human preference alignment with \acrshort{acr_dpo}.

\section{Evaluations \& Discussions}
We initially evaluate OpenBezoar-SFT and corresponding \acrshort{acr_qlora} adapter model on a set of standard benchmarks.
However, elevated scores against such standard benchmarks does not always comply with the requirement of appraising human preferences. In fact, the aligned model is demonstrated to have comparable scores to the base model for the aforementioned benchmarks. As we performed \acrshort{acr_sft} on OpenBezoar-SFT with the HH-RLHF dataset (conversation dataset as an instruction task) to obtain OpenBezoar-HH-RLHF-SFT, we were contented to evaluate OpenBezoar-HH-RLHF-DPO for human preferences alignment, as a chat assistant. Subsequently, we chose to utilize ``LLM-as-a-judge"\cite{zheng2023judging} framework, choosing ``claude-2.1"\cite{anthropic} as the judge, using the MT-bench benchmark. Furthermore, we evaluate our top performing model against MT-bench to compare against several purposefully chosen models in the 3B parameter scale in Open LLM Leaderboard\cite{huggingface-open-llm-leaderboard}.

\subsection{LM Eval Harness}
Given the vast number of different tasks in the LM Evaluation Harness \cite{EleutherAI} by EleutherAI, we narrowed down our focus to those that our selected base model has already been evaluated on\footnote{These original test results can be found at \href{https://huggingface.co/openlm-research/open_llama_3b_v2}{https://huggingface.co/openlm-research/open\_llama\_3b\_v2}}. We used the \texttt{big-refactor} branch\footnote{This branch can be found \href{https://github.com/EleutherAI/lm-evaluation-harness/tree/big-refactor}{https://github.com/EleutherAI/lm-evaluation-harness/tree/big-refactor}.} of the \texttt{lm-eval-harness} repo to conduct our evauation. After reproducing the same evaluations for the chosen tasks on the base model itself, we moved forward to evaluating our three checkpoints. These results, tabulated in Table \ref*{table:lm-eval-harness-results}, indicates how little improvement is observed in the first two checkpoints in comparison to the base model. Significant improvement over the base model is only observed in the DPO checkpoint by a value of 2\%.

\begin{table}[ht]
    \centering
    \begin{tabular}{p{0.11\linewidth}p{0.1\linewidth}p{0.155\linewidth}p{0.155\linewidth}p{0.155\linewidth}p{0.155\linewidth}}
        \toprule
        \parbox{0.11\linewidth}{\textbf{Task}} & \parbox{0.1\linewidth}{\textbf{Metric}} & \parbox{0.155\linewidth}{\textbf{OpenLLaMA 3B v2}} & \parbox{0.155\linewidth}{\textbf{OpenBezoar-SFT}} & \parbox{0.155\linewidth}{\textbf{OpenBezoar-HH-RLHF-SFT}} & \parbox{0.155\linewidth}{\textbf{OpenBezoar-HH-RLHF-DPO}} \\
        \midrule
        arc\_challenge                         & acc                                     & 0.3567                                             & 0.3720                                            & 0.3652                                                    & 0.3951                                                    \\
                                               & acc\_norm                               & 0.4036                                             & 0.4121                                            & 0.4044                                                    & 0.4309                                                    \\
        arc\_easy                              & acc                                     & 0.6991                                             & 0.7134                                            & 0.7117                                                    & 0.7336                                                    \\
                                               & acc\_norm                               & 0.7075                                             & 0.7088                                            & 0.7104                                                    & 0.7319                                                    \\
        hellaswag                              & acc                                     & 0.5278                                             & 0.5343                                            & 0.5254                                                    & 0.5580                                                    \\
                                               & acc\_norm                               & 0.7093                                             & 0.7077                                            & 0.6970                                                    & 0.7340                                                    \\
        mmlu                                   & acc                                     & 0.2648                                             & 0.2782                                            & 0.2675                                                    & 0.2614                                                    \\
                                               & acc\_norm                               &                                                    &                                                   &                                                           &                                                           \\
        openbookqa                             & acc                                     & 0.2940                                             & 0.3200                                            & 0.2900                                                    & 0.3380                                                    \\
                                               & acc\_norm                               & 0.4000                                             & 0.4200                                            & 0.4000                                                    & 0.4300                                                    \\
        piqa                                   & acc                                     & 0.7813                                             & 0.7807                                            & 0.7851                                                    & 0.7927                                                    \\
                                               & acc\_norm                               & 0.7889                                             & 0.7862                                            & 0.7889                                                    & 0.7982                                                    \\
        race                                   & acc                                     & 0.3895                                             & 0.3952                                            & 0.3828                                                    & 0.4239                                                    \\
                                               & acc\_norm                               &                                                    &                                                   &                                                           &                                                           \\
        sciq                                   & acc                                     & 0.9560                                             & 0.9520                                            & 0.9500                                                    & 0.9570                                                    \\
                                               & acc\_norm                               & 0.9580                                             & 0.9530                                            & 0.9580                                                    & 0.9600                                                    \\
        truthfulqa                             & acc                                     & 0.2581                                             & 0.2947                                            & 0.2669                                                    & 0.2866                                                    \\
                                               & acc\_norm                               &                                                    &                                                   &                                                           &                                                           \\
        winogrande                             & acc                                     & 0.6322                                             & 0.6338                                            & 0.6361                                                    & 0.6496                                                    \\
                                               & acc\_norm                               &                                                    &                                                   &                                                           &                                                           \\
        \midrule
        \textbf{Average}                       & \textbf{}                               & \textbf{0.5704}                                    & \textbf{0.5789}                                   & \textbf{0.5712}                                           & \textbf{0.5926}                                           \\
        \bottomrule                                                                                                                                                                                                                                                                                                       \\
    \end{tabular}
    \caption{LM Evaluation Harness results on all three of our checkpoints in comparison to the base model finetuned. Significant improvement is only observed in the DPO checkpoint. The average depicted in the table is the micro average across all the tasks.}
    \label{table:lm-eval-harness-results}
\end{table}

\subsection{Human Preferences Alignment}
As we anticipated on employing ``LLM-as-a-judge'' framework, we aimed to investigate ``judging'' capabilities of a mainstream model that offers free/discounted rates for academic research, as opposed to the OpenAI API. Recently Anthropic released their ``claude-2.1'' model boasting an impressive 200K context window and signficantly reduced rate of hallucinations when compared to its predecessor. Most importantly they allow early-access\footnote{Anthropic allows free usage with a capped request-limit} incrementally, based on the purpose of usage. However, in the context of models deployed by Anthropic, \cite{zheng2023judging} evaluates only ``claude-v1'' and their work focuses extensively only on the highest agreement with humans, which has patently been observed for GPT-4. Building on top of their work, we first established ``claude-2.1'' as a viable ``judge'' by computing and comparing agreement against other judges. Refer to the appendix \ref{appendix:establish_claude_2_1_judge} for a detailed explanation and results. We observed an impressive 88\% agreement level for Claude-2.1 with human majority votes when the ties are excluded. Remarkably, this surpasses the 85\% agreement of GPT-4 with the human majority.\par

Given that our fine-tuning approach emphasizes instruction-following abilities alongside conversational aptitude, MT-bench benchmark seamlessly aligns with our evaluation objectives\cite{zheng2023judging}. In contrast, Chatbot Arena do not rely specifically on restricted domains or use-cases, and therefore it lacks predefined questions. Furthermore, the HH-RLHF dataset we used for preference alignment contains instructions formulated as chat messages\cite{bai2022training}. On this basis, and given the nature of our training recipe, we concluded that it is sufficient to evaluate only against MT-bench.\par

In our experiments, we limit the maximum new token count in inference to 2048, which is higher than the limit of 1024 used in the evaluations in \cite{zheng2023judging}. This higher limit is based on the implicit need to penalize the models against undesired, meaningless repetitions in the response, where OpenBezoar-SFT was not just fine-tuned for open-end generation but for appropriate termination as well. Based on the need for a reference-guided judge, especially in math and reasoning questions, we initially prompted ``claude-2.1'' to obtain a reference answer to every question in the benchmark. We use single answer grading mode for the subsequent evaluations. There is no reason to refrain from using other modes if necessary but we leave it for an interested reader to pursue. We modified the implementation of the authors of \cite{zheng2023judging} by incorporating new models, including OpenBezoar-SFT\footnote{Modified Codebase: \href{https://bitbucket.org/paladinanalytics/fastchat}{https://bitbucket.org/paladinanalytics/fastchat}}.\par

\begin{figure}[ht]
    \centering
    \includegraphics[width=0.5\textwidth]{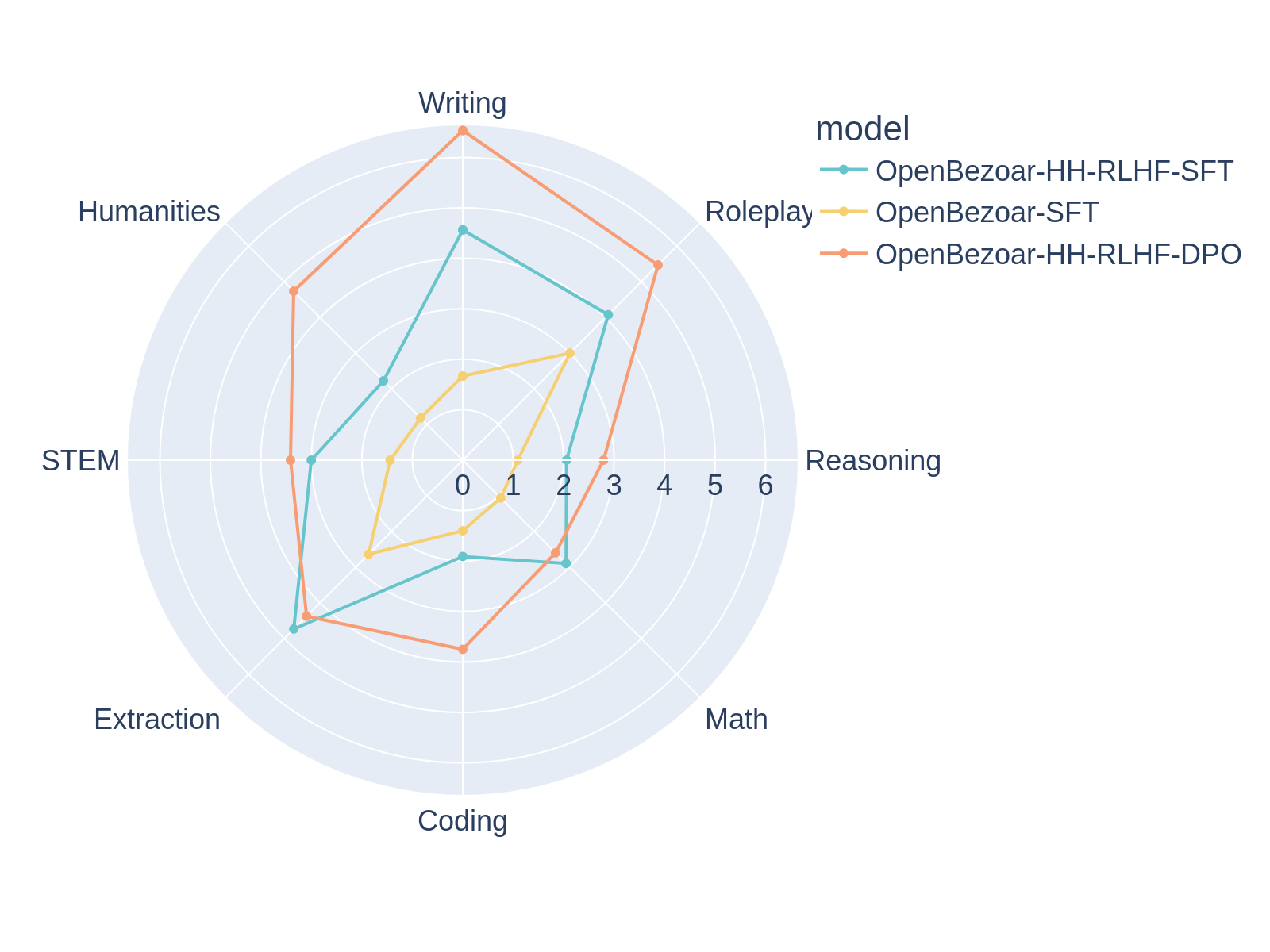}
    \caption{Category-wise scores for the OpenBezoar-SFT Models against MT-bench}
    \label{fig:mt_bench_scores_category_open_bezoar}
\end{figure}

Our initial attempt was to validate the DPO checkpoint as the best human preferences aligned model among OpenBezoar-SFT, OpenBezoar-HH-RLHF-SFT, and OpenBezoar-HH-RLHF-DPO models. Thus, we performed evaluations with each model to calculate the scores for each category w.r.t. first and second turns as described in \cite{zheng2023judging}, and finally obtained the average score. The overall average scores for each model are given in the Table \ref{table:mt_bench_scores_average_open_bezoar}. By plotting the average score for each category, we obtain the radar plot given in the Figure \ref{fig:mt_bench_scores_category_open_bezoar}. On average, OpenBezoar-HH-RLHF-SFT has seen a drastic improvement over OpenBezoar-SFT. This must be largely due to the HH-RLHF dataset subset size. However, the second turn score has not exhibited a comparable scale of improvement to that seen in the first turn. Nevertheless, OpenBezoar-HH-RLHF-DPO has achieved the best average score and scores for each turn. Notably, the improvement over the second turn when compared to OpenBezoar-HH-RLHF-SFT has significantly been higher for OpenBezoar-HH-RLHF-DPO. While a comprehensive understanding of the distribution of the HH-RLHF dataset is necessary to provide a definitive explanation, it is plausible that the improvement in the second turn scores in OpenBezoar-HH-RLHF-DPO could be attributed to the methodology behind fine-tuning to derive OpenBezoar-HH-RLHF-SFT and lack of dispreferred responses among the generated first turn responses during inference (using the preference-aligned model). Recall that OpenBezoar-HH-RLHF-SFT was fine-tuned on the ``chosen'' responses from the dataset and HH-RLHF may contain dispreferred responses within the preceding turns, even among the ``chosen'' responses. Furthermore, it is noticeable that OpenBezoar-HH-RLHF-DPO slightly underperforms OpenBezoar-HH-RLHF-SFT in extraction and math categories. Nevertheless, we conclude that OpenBezoar-HH-RLHF-DPO stands out as the best-performing model to emerge from our training recipe.\par

\begin{table}[ht]
    \centering
    \begin{tabular}{cccc}
        \toprule
        Model                  & First Turn & Second Turn & Average \\
        \midrule
        OpenBezoar-SFT         & 1.82       & 1.57        & 1.68    \\
        OpenBezoar-HH-RLHF-SFT & 4.11       & 2.47        & 3.23    \\
        OpenBezoar-HH-RLHF-DPO & 4.79       & 3.44        & 4.12    \\
        \bottomrule
    \end{tabular}
    \caption{Average Scores of the OpenBezoar-SFT models against MT-bench. First Turn and Second Turn refers to the number of turns each party has taken in the conversation, as defined in \cite{zheng2023judging}. The scores have been rounded off to the nearest second decimal place.}
    \label{table:mt_bench_scores_average_open_bezoar}
\end{table}

\begin{figure}[ht]
    \centering
    \includegraphics[width=0.5\textwidth]{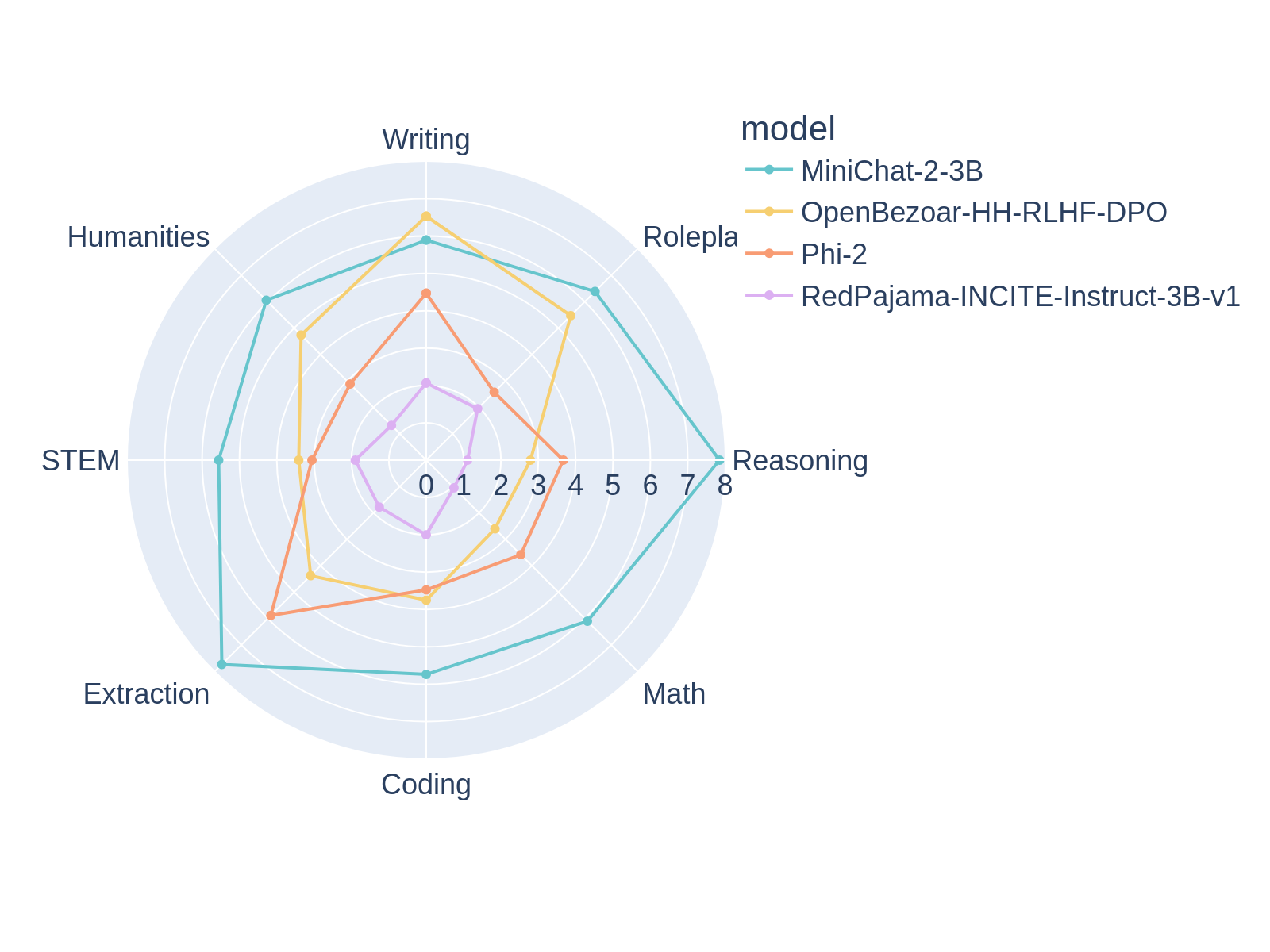}
    \caption{Category-wise scores for a set of chosen Models from Open LLM Leaderboard against MT-bench}
    \label{fig:mt_bench_scores_category_all_models}
\end{figure}

To assess our top-performing model against others, we first needed to identify other candidate models for comparison. Our motivation during the initial stages was to assess the instruction following and conversational capabilities of 3B scale \acrshort{acr_llm}s. To the best of our knowledge, ``RedPajama-INCITE-Chat-3B-v1'' was the sole well-documented model in 3B scale for conversational capabilities during the time we started our experiments. Consequently, we opted for this model as our first candidate for comparison against OpenBezoar-HH-RLHF-DPO. Additionally, we expected to report the performance of a properly documented pre-eminent 3B scale model in the Open LLM Leaderboard\cite{huggingface-open-llm-leaderboard}, that has also fine tuned for conversation against our top-performer. Examining and filtering the models in the leaderboard for our criteria, we selected ``MiniChat-2'' family\cite{zhang2023law} and more specifically ``MiniChat-2-3B'' model as the next candidate. Despite its prior evaluation with MT-bench, our selection for the judge differs from theirs. Therefore, we proceeded to recalculate the scores using "claude-2.1" as the judge. Lastly, with the intention of evaluating against a model with slightly less parameters count, we chose ``Phi-2'' more or less arbitrarily. Although it is announced as a base model, it supports chat completions and it is reported as one of the intended use cases.
Similar to the previous evaluation pipeline, we report the overall average score in the Table \ref{table:mt_bench_scores_average_all_models} followed up by a category-wise plot of the scores in the Figure \ref{fig:mt_bench_scores_category_all_models}. It is evident that OpenBezoar-HH-RLHF-DPO ranks second among the chosen models in terms of the average score. Most importantly, it surpasses ``RedPajama-INCITE-Chat-3B-v1'', the sole 3B scale model at the time of initial experimentation, with a significant margin. With the exception of three categories, our OpenBezoar-HH-RLHF-DPO outstrips Phi-2, which is noteworthy considering that Phi-2, despite being designated as a base model, has been trained on larger datasets and fine-tuned for chat completion. However, it should be noted that Phi-2 itself has apparently not been trained for generating the End-Of-Stream(EOS) token, i.e., to terminate the response when appropriate. Hence, repetitions were observed in the model responses and has had adversely affected on the scores. However, we refrain from applying any form of generation control to ensure a fair evaluation scheme, even though doing so might have led to improved scores for Phi-2. As previously stated, ``MiniChat-2-3B'' is one of the best-performers in the 3B scale models and the scores are self-justifiable, given their training scheme with a better mixture of data. Nevertheless, OpenBezoar-HH-RLHF-DPO has outperformed ``MiniChat-2-3B'' in the category of ``Writing'', which we ascribe to HH-RLHF dataset distribution. Therefore, it seems evident that a better mixture of data might be useful rather than just a large dataset, even for human preference alignment with \acrshort{acr_dpo}. Furthermore, given that we only conducted \acrshort{acr_dpo} for a single epoch, it is advisable to train for several additional epochs.\par

\begin{table}[ht]
    \centering
    \begin{tabular}{cccc}
        \toprule
        Model                       & First Turn & Second Turn & Average \\
        \midrule
        OpenBezoar-HH-RLHF-DPO      & 4.79       & 3.44        & 4.12    \\
        RedPajama-INCITE-Chat-3B-v1 & 1.57       & 1.33        & 1.45    \\
        MiniChat-2-3B               & 6.87       & 6.00        & 6.43    \\
        Phi-2                       & 4.43       & 2.99        & 3.72    \\
        \bottomrule
    \end{tabular}
    \caption{Average Scores of a set of chosen Models against MT-bench. First Turn and Second Turn refers to the number of turns each party has taken in the conversation, as defined in \cite{zheng2023judging}. The scores have been rounded off to the nearest second decimal place.}
    \label{table:mt_bench_scores_average_all_models}
\end{table}

\paragraph{Experimental Setup} As specified earlier, we modified the official implementation of \cite{zheng2023judging} to accommodate the changes that are required to evaluate our models. Utilizing this revised codebase, we leveraged Kaggle's free quota, specifically a Tesla P100 GPU, for generating model responses and conducting evaluations. Please refer to the ReadMe in the codebase under "fastchat/llm\_judge" for detailed instructions. Subsequently, we utilized a CPU runtime on either Kaggle or Google Colaboratory for score calculations and generating category-wise radar plots.\par

\section{Conclusion}
Our study focused primarily on generating synthetic data for instruction following on datasets with a few prominent schemes and an open model, and using this data to fine-tune a small open base model to establish this use case, producing the OpenBezoar family of models. Open models available at the time of this work that permitted commercial use were utilized throughout except for the filtering of these datasets and the evaluation of the models.\par

This work also explored the impact of \acrshort{acr_sft} with adapters and in particular \acrshort{acr_qlora} in this setting, further minimizing compute costs, as well as fine-tuning for alignment with \acrshort{acr_dpo}. The resulting checkpoints were evaluated on a set of ten benchmarks from LM-Eval-Harness and with single answer grading on MT-Bench. On the former, on almost all benchmarks we saw significant improvements with ``OpenBezoar-SFT'' and ``OpenBezoar-HH-RLHF-SFT'' over the base model and similar improvements with ``OpenBezoar-HH-RLHF-DPO'' over ``OpenBezoar-SFT''. On the latter, our results indicate that there is a significant improvement in ``OpenBezoar-HH-RLHF-SFT'' over ``OpenBezoar-SFT'' in every task category and in the average score. ``OpenBezoar-HH-RLHF-DPO'' exhibits varying degrees of improvement over ``OpenBezoar-HH-RLHF-SFT'' in different tasks, however, with the exception of a minor degradation of the score in two task categories (Math and Extraction).\par

We note some key limitations of the current work and propose some directions for further work.\par

\subsection{Limitations and Future Directions}

\begin{itemize}
    \item Given the limitations of open models available at the time of this work, the synthetic data generated from the chosen parent model exhibits some irregular characteristics. There were instances where the parent model did not produce the desired output as outlined in the prompt. Stronger open models and more crowd-sourced open instruction datasets are emerging that may be utilized to address this gap.
    \item Our datasets consisted of a relatively small number of examples, similar to the study conducted by \citeauthor{zhou2024lima} \cite{zhou2024lima}. However, unlike their study, we did not curate the examples meticulously for fine-tuning our base model except for filtering the generations with GPT-4. This, coupled with the fact that most capable instruction-tuned models on the 3B parameter scale  are fine-tuned on considerably larger datasets, may contribute to ``OpenBezoar-SFT's'' relatively minor improvement over its base model. More sophisticated in-context learning and agentic patterns may be utilized to implement automatic curation schemes during or post-generation to mitigate this limitation.
    \item The use of GPT-4 for filtering generations and the use of Claude-2.1 for evaluation mean we have not fully extricated ourselves from the restrictions imposed by closed-source services. Fine-tuning an open base model for this kind of critique/evaluation is a valuable direction for future work.
    \item It should be noted that the models released from this study might not exhibit enhanced instruction-following capabilities. There could be instances where the models respond out of context or fall into a pattern of looping generations, leading to the repetition of a word or a group of words. More diversity in the types of instruction tasks present in the instruction schemes, as well as different mixes of instructions from different schemes used for fine-tuning (compared to the sequential fine-tuning done here) are worthy of further exploration.
    \item The inclusion of the system prompt is crucial when executing generations with the fine-tuned models. Without it, the generated responses may appear nonsensical. Proper responses can only be produced if the system prompt is one that the model was fine-tuned on. Whether a model can be made to learn over different system prompts that convey the same meaning and then respect out of domain system prompts at inference time is a key question to be investigated.
    \item The full \acrshort{acr_sft} on the HH-RLHF dataset was done after merging the LoRA weights with the base model on ``OpenBezoar-SFT''. More sophisticated merging methods may be explored here. Another exploration possible here is to evaluate and compare the performance of applying \acrshort{acr_dpo} with a low-rank adapter instead of the merged model.
    \item While we release our strongest ``OpenBezoar-HH-RLHF-DPO'' checkpoint fine-tuned for alignment with human preferences, it may still diverge from these in unexpected ways. The further constraints under which the entire family of OpenBezoar models have been trained as described in this paper urge caution as to what uses they ought to be put to: in particular, we caution against reliance on them for production or adjacent use-cases where robust responses are required and where hallucinations, bias, toxicity, and general divergence from intended application are not acceptable.
\end{itemize}

\pagebreak

\begin{appendices}

    \section{LaMini Prompts}
    \label{appendix:lamini_prompts}
    \begin{figure}[H]
    \centering
    \fbox{
        \begin{minipage}{\linewidth}
            \#\#\# SYSTEM: You are an AI assistant. Answer as honestly and correctly as possible.\\
            \#\#\# YOUR TASK: Generate 5 diverse examples that are similar to the provided examples.\\
            You do not need to provide responses to the generated examples.\\
            Do not repeat the provided examples.\\
            Each generated example must include an instruction.\\
            Each generated example may have an additional context if necessary.\\
            Each generated example can be either an imperative sentence or a question.\\
            Each generated example must begin with "<example>" and end with "</example>"\\
            Each generated example should be themed on one of the topics of American philosophers,Hume Highway,Finance ministries\\

            \#\#\# PROVIDED
            EXAMPLES(Category: closed\_qa):\\
            <example
            >What is linux Bootloader\\ Input:A bootloader, also spelled as boot loader or
            \dots</example
            >\\
            <example
            >What is one-child policy?\\ Input:The term one-child policy refers to a
            population planning initiative in \dots</example
            >\\
            <example
            >When was Tomoaki Komorida born?\\ Input:Komorida was born in Kumamoto
            Prefecture on July 10, 1981. After \dots</example
            >\\
            \#\#\#RESPONSE:
        \end{minipage}
    }
    \caption{A topic guided prompt used for
        creating datasets with the LaMini scheme from the \texttt{h2oai/h2ogpt-gm-oasst1-en-2048-falcon-40b-v2} parent model. The three dots(\ldots) shown at the end of each example in this prompt is only depcited as a truncation of the original example.}
    \label{fig:lamini_topic_guided_prompt} \end{figure}

    \section{Evol-Instruct Prompts}
    \label{appendix:evol_instruct_prompts}
    \begin{figure}[H]
    \centering
    \fbox{
        \begin{minipage}{\linewidth}
            <human>: I want you to act as a prompt rewriter.\\
            Your objective is to rewrite the \#Given Prompt\# into a more complex version.\\
            But the rewritten prompt must be reasonable and must be understood and responded by humans.\\
            Your rewriting cannot omit the non-text parts such as the table and code in \#Given Prompt\#:. Also, please do not omit the context in \#Given Prompt\#.\\
            You should try your best not to make the \#Rewritten Prompt\# become verbose, \#Rewritten Prompt\# can only add 10 to 20 words into \#Given Prompt\#.\\
            '\#Given Prompt\#', '\#Rewritten Prompt\#', 'given prompt' and 'rewritten prompt' are not allowed to appear in \#Rewritten Prompt\#\\
            You SHOULD complicate the given prompt if \#Given Prompt\# contains inquiries about certain issues, the depth and breadth of the inquiry can be increased.\\
            \#Given Prompt\#:\\
            Why did Syd Barrett left the Pink Floyd?\\
            <bot>: \#Rewritten Prompt\#:
        \end{minipage}
    }
    \caption{In depth evolving prompt for deepening a given instruction}
    \label{fig:evol_in_depth_evolving_deepening}
\end{figure}

\begin{figure}[H]
    \centering
    \fbox{
        \begin{minipage}{\linewidth}
            <human>: I want you to act as a prompt rewriter.\\
            Your objective is to rewrite the \#Given Prompt\# into a more complex version.\\
            But the rewritten prompt must be reasonable and must be understood and responded by humans.\\
            Your rewriting cannot omit the non-text parts such as the table and code in \#Given Prompt\#:. Also, please do not omit the context in \#Given Prompt\#.\\
            You should try your best not to make the \#Rewritten Prompt\# become verbose, \#Rewritten Prompt\# can only add 10 to 20 words into \#Given Prompt\#.\\
            '\#Given Prompt\#', '\#Rewritten Prompt\#', 'given prompt' and 'rewritten prompt' are not allowed to appear in \#Rewritten Prompt\#\\
            You SHOULD complicate the given prompt by replacing general concepts with more specific concepts.\\
            \#Given Prompt\#:\\
            Why did Syd Barrett left the Pink Floyd?\\
            <bot>: \#Rewritten Prompt\#:
        \end{minipage}
    }
    \caption{In depth evolving prompt for concretizing a given instruction}
    \label{fig:evol_in_depth_evolving_concretizing}
\end{figure}

\begin{figure}[H]
    \centering
    \fbox{
        \begin{minipage}{\linewidth}
            <human>: I want you to act as a prompt rewriter.\\
            Your objective is to rewrite the \#Given Prompt\# into a more complex version.\\
            But the rewritten prompt must be reasonable and must be understood and responded by humans.\\
            Your rewriting cannot omit the non-text parts such as the table and code in \#Given Prompt\#:. Also, please do not omit the context in \#Given Prompt\#.\\
            You should try your best not to make the \#Rewritten Prompt\# become verbose, \#Rewritten Prompt\# can only add 10 to 20 words into \#Given Prompt\#.\\
            '\#Given Prompt\#', '\#Rewritten Prompt\#', 'given prompt' and 'rewritten prompt' are not allowed to appear in \#Rewritten Prompt\#\\
            You SHOULD complicate the given prompt if \#Given Prompt\# can be solved with just a few simple thinking processes, you can rewrite it to explicitly request multiple-step reasoning.\\
            \#Given Prompt\#:\\
            Why did Syd Barrett left the Pink Floyd?\\
            <bot>: \#Rewritten Prompt\#:
        \end{minipage}
    }
    \caption{In depth evolving prompt for increasing reasoning steps in a given instruction}
    \label{fig:evol_in_depth_evolving_reasoning}
\end{figure}

    \section{Loss Charts during Q-LoRA Finetuning}
    \label{appendix:q_lora_finetuning_charts}
    \begin{figure}[H]
    \centering
    \begin{subfigure}{0.45\textwidth}
        \includegraphics[width=\linewidth]{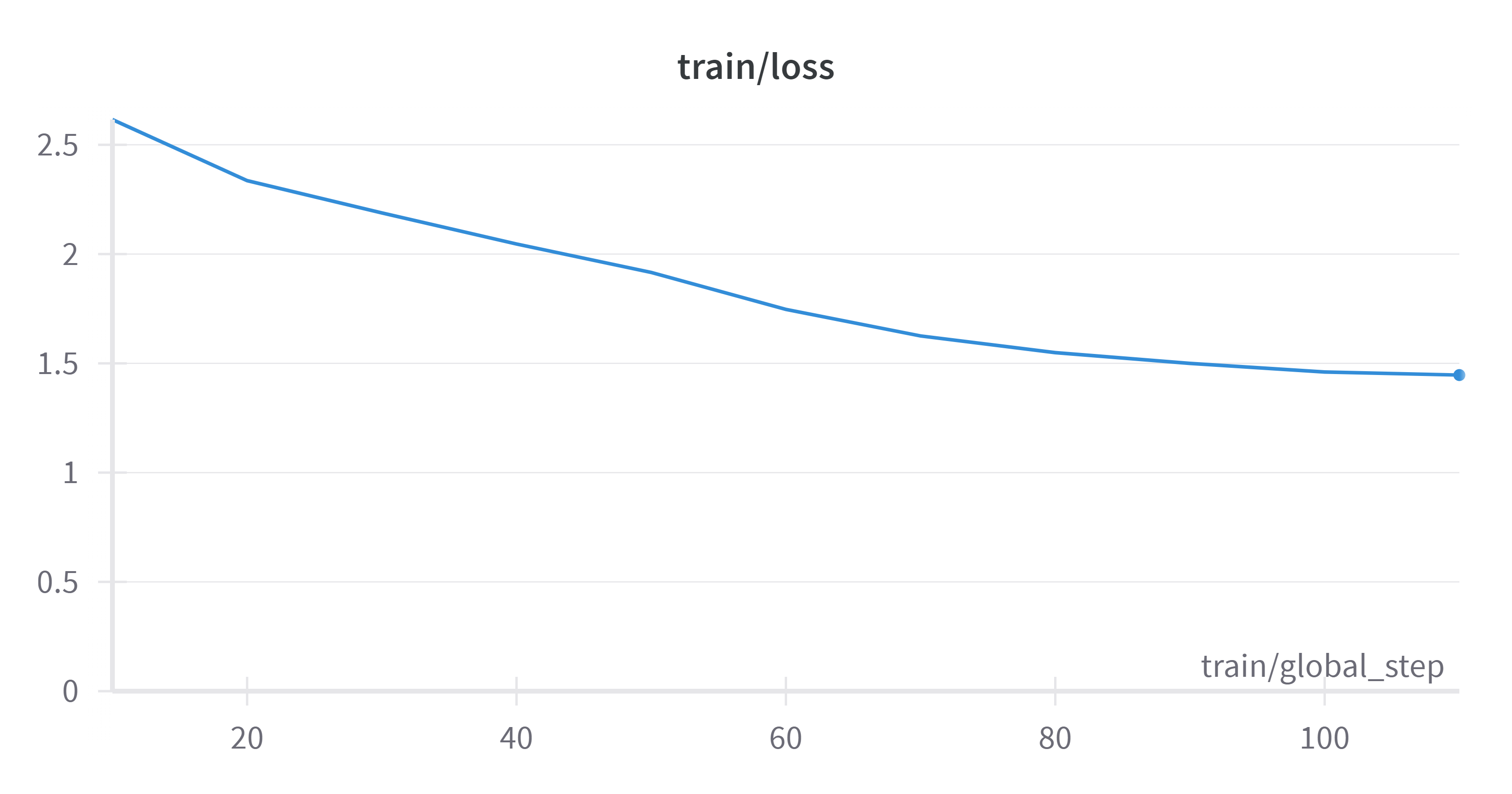}
        \caption{Train Loss during Run \#1}
    \end{subfigure}
    \begin{subfigure}{0.45\textwidth}
        \includegraphics[width=\linewidth]{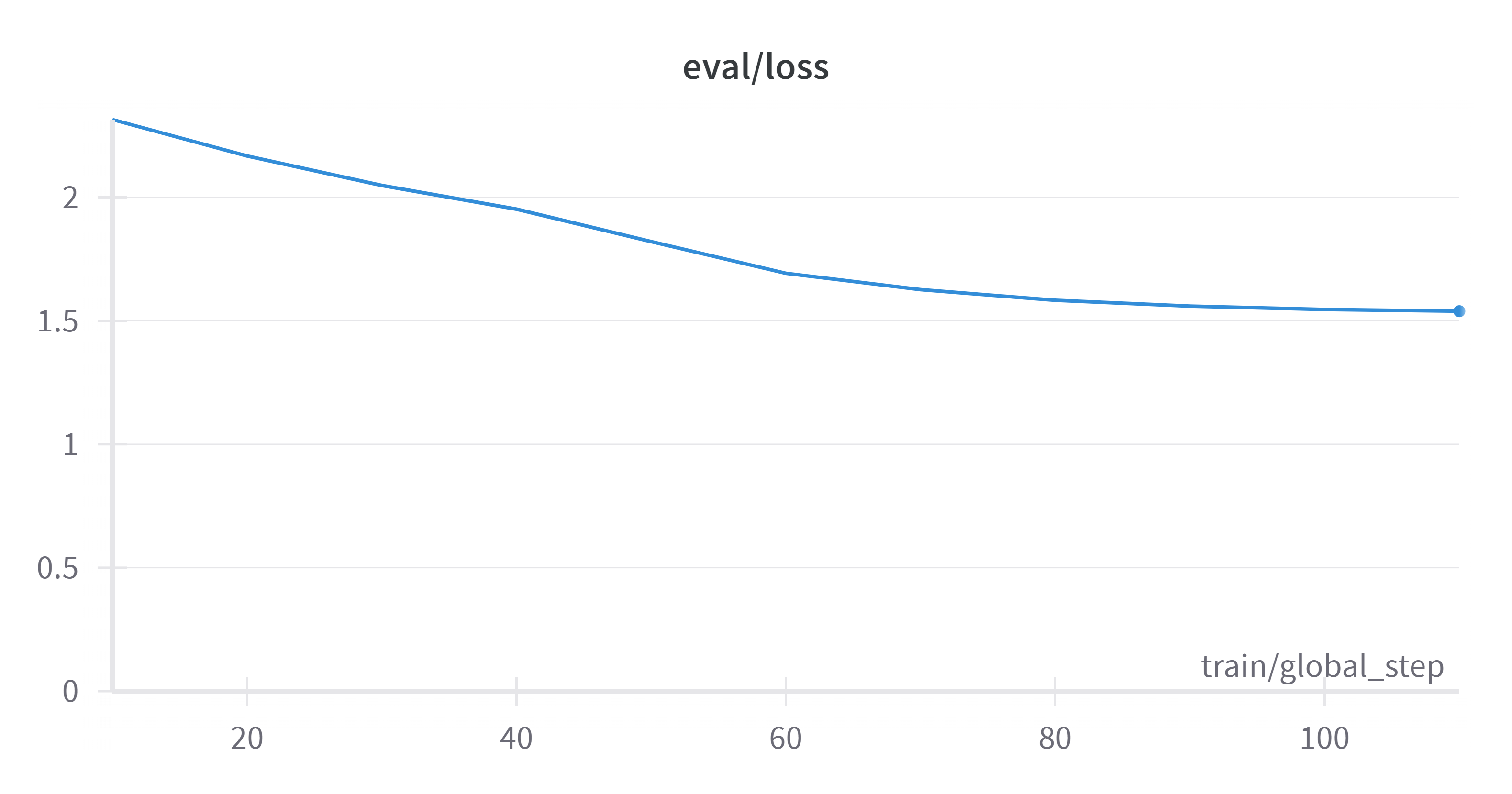}
        \caption{Eval Loss during Run \#1}
    \end{subfigure}

    \vspace*{2em}

    \begin{subfigure}{0.45\textwidth}
        \includegraphics[width=\linewidth]{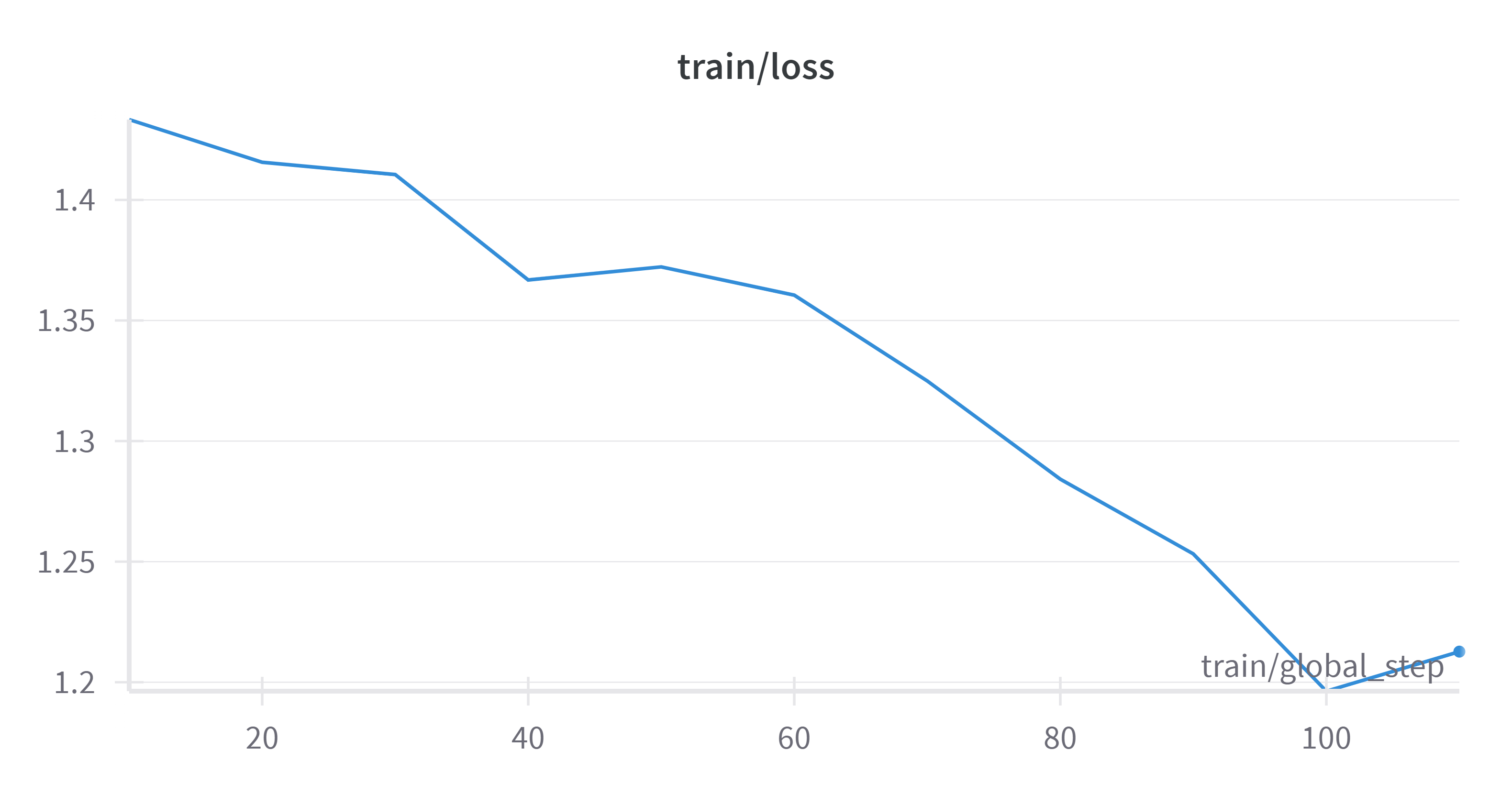}
        \caption{Train Loss during Run \#2}
    \end{subfigure}
    \begin{subfigure}{0.45\textwidth}
        \includegraphics[width=\linewidth]{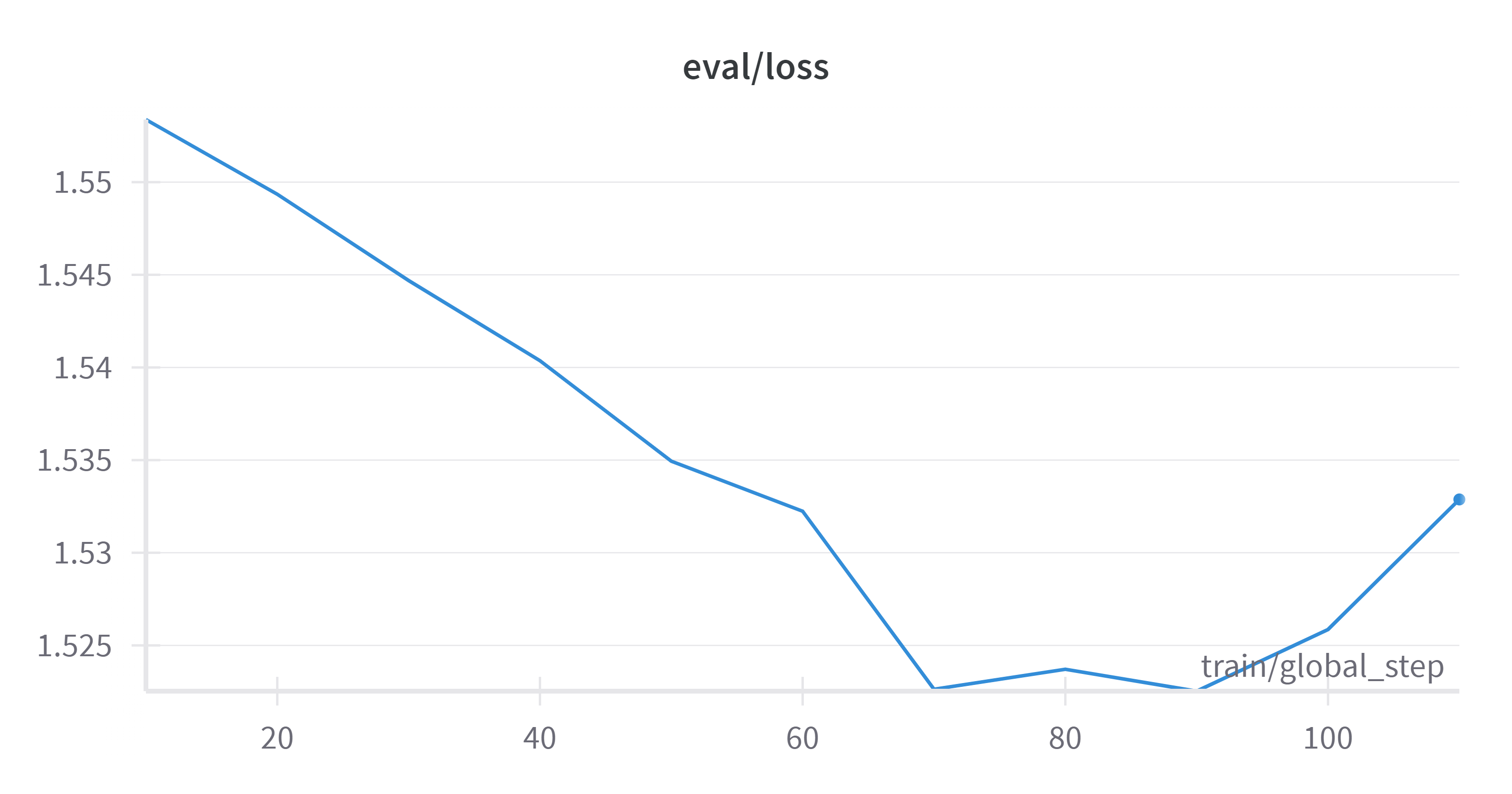}
        \caption{Eval Loss during Run \#2}
    \end{subfigure}

    \caption{Train and Eval loss during Orca finetuning}
    \label{fig:orca-loss-plots}
\end{figure}

\begin{figure}[H]
    \centering
    \begin{subfigure}{0.45\textwidth}
        \includegraphics[width=\linewidth]{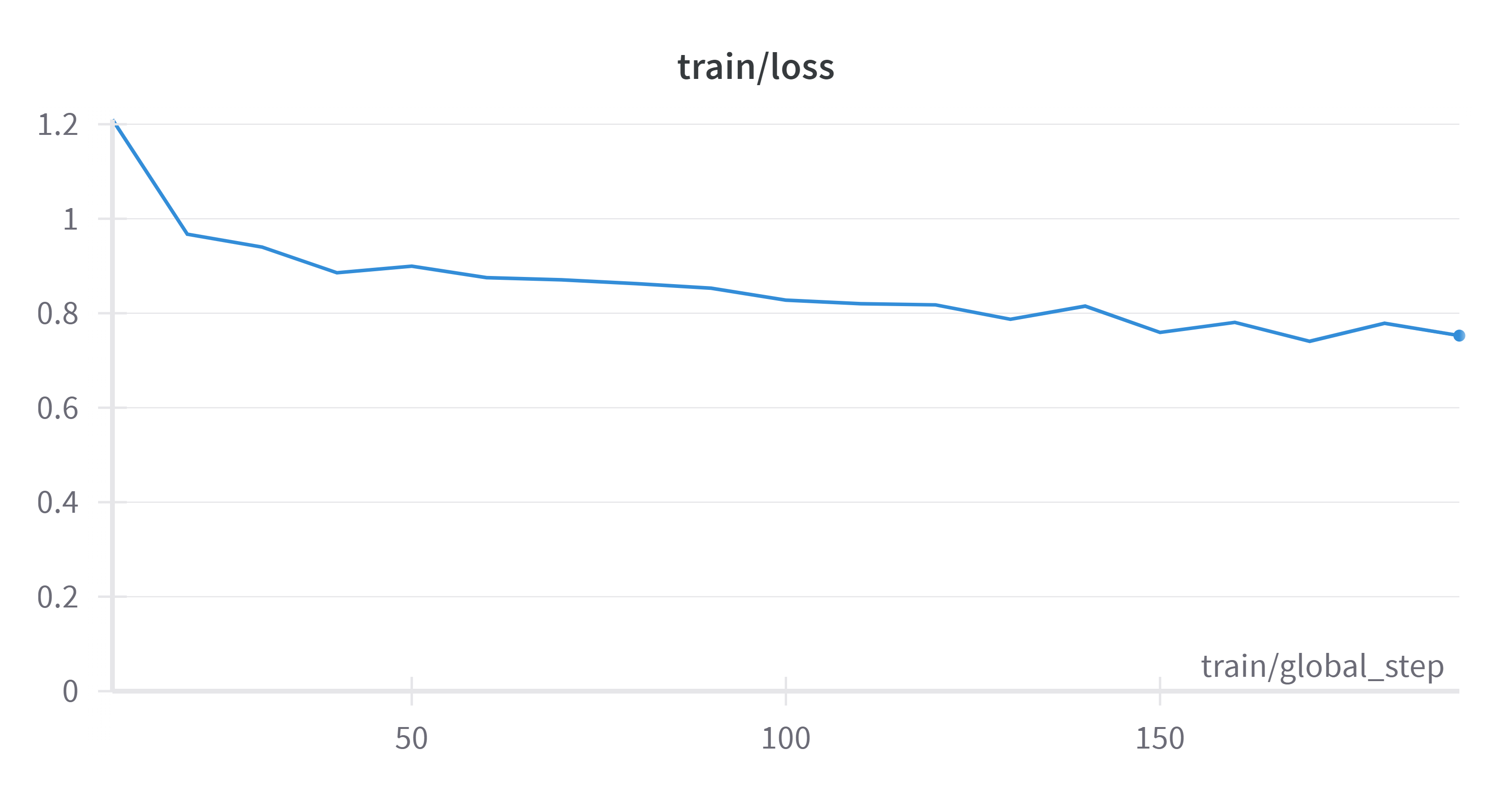}
        \caption{Train Loss during Run \#1}
    \end{subfigure}
    \begin{subfigure}{0.45\textwidth}
        \includegraphics[width=\linewidth]{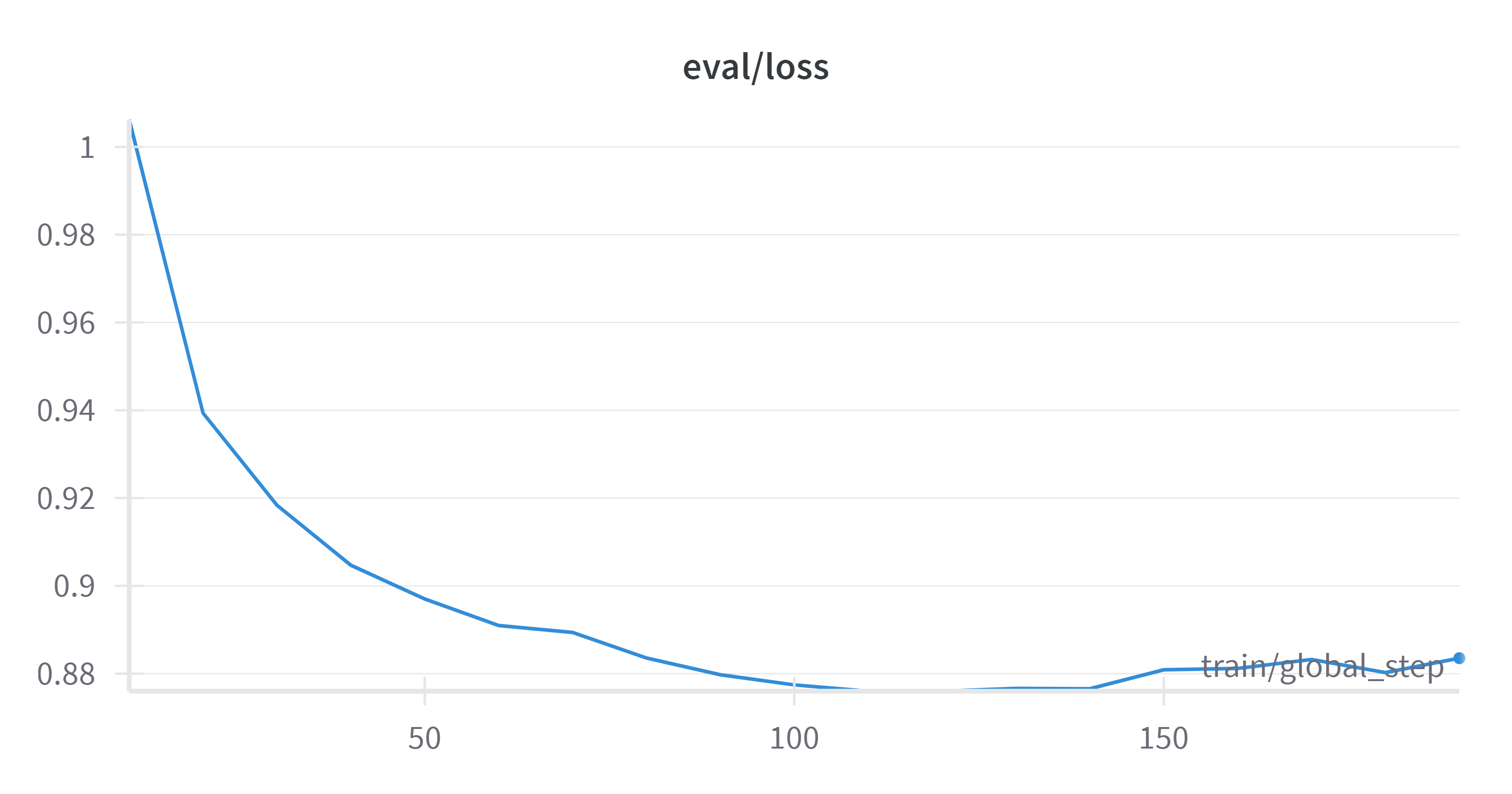}
        \caption{Eval Loss during Run \#1}
    \end{subfigure}

    \caption{Train and Eval loss during Evol-Instruct finetuning}
    \label{fig:evol-loss-plots}
\end{figure}

    \section{Establishing Claude-2.1 as a Judge}
    \label{appendix:establish_claude_2_1_judge}
    Our prime objective in this regard is to evaluate the agreement of Claude-2.1 as a judge, following a similar approach to \cite{zheng2023judging}. This necessitates the need for many expensive processes and resources such as human expert labelers and a dataset which contains pairwise responses to the questions in the benchmarks. However, in order to utilize the minimal resources we possessed for experimentation, we agreed upon several presets regarding the experimental setup. Similar to \cite{zheng2023judging}, we compute the agreement over the questions in MT-bench dataset between claude-2.1 and 5 judges: G4-S, G4-P, Author, Human, and Human-M separately, where the abbreviated terms are defined identical to the definitions in \cite{zheng2023judging}. Moreover, our evaluations will separately consider evaluations that include ties (S1) and those that exclude ties (S2). In contrast to the agreement evaluations w.r.t claude-v1 in \cite{zheng2023judging}, we also conduct the evaluations for the second turn with claude-2.1. However, our extreme resource constraints prevented us from evaluating claude-2.1 for limitations such as positional bias, verbosity bias etc.\par

To the best of our knowledge, the judgements for Author, Human, and Human-M judges were not publicly available for the MT-bench questions. In fact, only G4-S and G4-P judgements were available at the Huggingface MT-bench Leaderboard\cite{lmsys_mt-bench}. Furthermore, it was also noted that single-answer grading was converted to the pairwise results during the agreement evaluation. Therefore, we initially converted the G4-S judgements found in \cite{lmsys_mt-bench} to the pairwise form by pairing up every possible answer pair for a given question and considering the model corresponding to the answer with the highest score as the winner. The resulting judgements are deployed at: \href{https://huggingface.co/datasets/chansurgeplus/mt_bench_gpt4_single_pairs_judgments}{https://huggingface.co/datasets/chansurgeplus/mt\_bench\_gpt4\_single\_pairs\_judgments}. However, when compared to the MT-bench human judgements, publicly available at \href{https://huggingface.co/datasets/lmsys/mt_bench_human_judgments}{https://huggingface.co/datasets/lmsys/mt\_bench\_human\_judgments}, not every response pair in our ``mt\_bench\_gpt4\_single\_pairs\_judgments'' judgements dataset has been human evaluated. Furthermore, the corresponding G4-P evaluations had an even smaller subset of judgments. Subsequently, in order to ensure a fair evaluation scheme for every judge considered, we derived the intersection of all 3 judgement datasets: ``mt\_bench\_gpt4\_single\_pairs\_judgments'', ``mt\_bench\_human\_judgments'' and its corresponding G4-P judgements. This resulted in a judgements dataset with 640 records, published at \href{https://huggingface.co/datasets/chansurgeplus/mt_bench_gpt4_single_pairs_overlap_judgments}{https://huggingface.co/datasets/chansurgeplus/mt\_bench\_gpt4\_single\_pairs\_overlap\_judgments}. This eliminates the bias due to the sample size in our evaluations.\par

%% Then we evaluate the pairs in that dataset using claude on different presets.
Next, we allowed claude-2.1 to generate judgements on the response pairs in the ``mt\_bench\_gpt4\_single\_pairs\_overlap\_judgments'' dataset. The temperature was fixed at 0 for all generations and the maximum length of generation was limited to 1024. Using the resulting judgements, the agreement ratios were computed and are reported in Table \ref{table:appendix_claude_estb_results}.\par

\begin{table}[ht]
    \centering
    \begin{subtable}[b]{1\linewidth}
        \begin{tabular}{l*{11}{c}}
            \toprule
            Setup                       & \multicolumn{5}{c}{S1} & \multicolumn{5}{c}{S2}                                                                                                                       \\ \cmidrule(lr){2-6} \cmidrule(lr){7-11}
            Judge                       & G4-S                   & G4-P                   & Author      & Human        & Human-M      & G4-S         & G4-P         & Author      & Human        & Human-M      \\
            \midrule
            \multirow{2}{*}{Claude-2.1} & 71\%                   & 70\%                   & 57\%        & 61\%         & 61\%         & 96\%         & 98\%         & 92\%        & 91\%         & 88\%         \\
                                        & \deemph{224}           & \deemph{222}           & \deemph{63} & \deemph{217} & \deemph{334} & \deemph{189} & \deemph{194} & \deemph{54} & \deemph{184} & \deemph{308} \\
            \bottomrule
        \end{tabular}
        \caption{First Turn}
        \label{table:appendix_claude_estb_results_turn_1}
    \end{subtable}%

    \begin{subtable}[b]{1\linewidth}
        \begin{tabular}{l*{11}{c}}
            \toprule
            Setup                       & \multicolumn{5}{c}{S1} & \multicolumn{5}{c}{S2}                                                                                                                       \\ \cmidrule(lr){2-6} \cmidrule(lr){7-11}
            Judge                       & G4-S                   & G4-P                   & Author      & Human        & Human-M      & G4-S         & G4-P         & Author      & Human        & Human-M      \\
            \midrule
            \multirow{2}{*}{Claude-2.1} & 67\%                   & 60\%                   & 62\%        & 55\%         & 55\%         & 95\%         & 96\%         & 92\%        & 89\%         & 85\%         \\
                                        & \deemph{198}           & \deemph{177}           & \deemph{61} & \deemph{185} & \deemph{295} & \deemph{166} & \deemph{161} & \deemph{47} & \deemph{156} & \deemph{259} \\
            \bottomrule
        \end{tabular}
        \caption{Second Turn}
        \label{table:appendix_claude_estb_results_turn_2}
    \end{subtable}%

    \caption{Agreement between Claude-2.1 and pre-determined judges, evaluated against MT-bench. To quote \cite{zheng2023judging}, ``G4-S'', ``G4-P'', and ``Human-M'' denote GPT-4 with pairwise comparison, GPT-4 with single answer grading, and majority vote of humans respectively. Author refers to the human who authored the question. The two setups ``S1'' and ``S2'' are exactly the same as defined in \cite{zheng2023judging}. Accordingly, ``S1'' includes tie votes, possibly due to positional bias inconsistencies whereas ``S2'' does not. The top value in each cell represents the agreement percentage (calculated against the total number of response pairs in the subset for respective setup) and the bottom greyed value denotes the agreed votes count.}
    \label{table:appendix_claude_estb_results}
\end{table}

\paragraph{Experimental Setup} We utilized free, non-accelerated Kaggle notebooks for generating judgements. Based on the allowed request rate to the Anthropic API, single-turn and two-turn judgement generations were conducted simultaneously. Outputs were logged internally. Both evaluations were completed nearly after 2 hours.\par

\paragraph{Results} We observe a high agreement of Claude-2.1 with human and human-majority votes. It is important to emphasize that Claude-2.1 agreement with humans has slightly transcended ``G4-S'' agreement with humans by 1\%, but this may very well be due to the lesser number of response pairs in our judgements dataset. Although ``G4-P'' surpasses Claude-2.1 by 6\% in ``S1'', Claude-2.1 has outperformed ``G4-S'' and ``G4-P'' agreement with human votes in first (single) turn evaluations in ``S2'', i.e., when ties are excluded. In fact, the agreement between Claude-2.1 and human majority is 88\% (w/o ties), which is further higher than both GPT-4 agreements with humans and agreement among humans, reported in \cite{zheng2023judging}. This holds true for the second turn as well, although the agreement has decreased by 3\%. Therefore, the agreement levels observed for Claude-2.1 with human experts are the highest for any judge so far(other than ``Human'' and ``Human-M'' is ignored, based on the consensus that we are targeting for LLM judges), whenever the ties are excluded. It can also be observed that there is an improvement of the agreement of Claude-2.1 with humans when compared to claude-v1 in the first turn. As \cite{zheng2023judging} does not report the agreement of claude-v1 for the second turn, we cannot comment on the improvement in that regard. Nevertheless, it can be fairly concluded that Claude-2.1 is a viable judge and can be used in evaluating responses of chat assistants.\par

\paragraph{Future Work} A breakdown analysis similar to \cite{zheng2023judging} is beyond the scope of this work and could be of interest for an extesive agreement and limitations evaluation of Claude-2.1 as a judge. Moreover, an enthusiastic individual might be interested in exploring the agreement evaluation of single-answer grading with Claude-2.1 as the judge. Since our primary goal was to demonstrate Claude-2.1 as a feasible judge, we leave the exploration of such pursuits for future endeavors. It is also a worthwhile experiment to validate the agreement levels observed for Claude-2.1 with a larger judgements dataset, provided there are enough ``human expert'' resources available to label and annotate the dataset.\par

\end{appendices}

\clearpage
\printbibliography

\end{document}